\documentclass[letterpaper, 10 pt, conference]{ieeeconf}  

\makeatletter
\let\NAT@parse\undefined
\makeatother

\IEEEoverridecommandlockouts                              

\usepackage{graphicx} 
\usepackage[numbers]{natbib}
\usepackage{amsmath} 
\usepackage{amssymb}  
\usepackage{mathtools}
\usepackage[font={small}]{caption}
\usepackage[]{algorithm2e}
\usepackage{subfig}
\usepackage{color}

\newcommand{\figref}[1]{Fig.~\ref{#1}}
\newcommand{\secref}[1]{Section~\ref{#1}}

\newcommand{\PAR}[1]{\vskip3pt \noindent{\bf #1~}}

\newcommand{\zh}[1]{\noindent{\color{black}{#1}}}
\newcommand{\fvzh}[1]{\noindent{\color{black}{#1}}}

\title{\LARGE \bf
Real-Time Dense Mapping \\for Self-Driving Vehicles using Fisheye Cameras}

\author{
\authorblockN{Zhaopeng Cui$^{1}$, Lionel Heng$^{2}$, Ye Chuan Yeo$^{2}$, Andreas Geiger$^{3}$, Marc Pollefeys$^{1,4}$, and Torsten Sattler$^{5}$}
\thanks{$^{1}$Department of Computer Science, ETH Z\"{u}rich}%
\thanks{$^{2}$DSO National Laboratories}%
\thanks{$^{3}$MPI-IS and University of T\"{u}bingen}%
\thanks{$^{4}$Microsoft, Switzerland}%
\thanks{$^{5}$Chalmers University of Technology}
}%

\begin{document}

\maketitle
\thispagestyle{empty}
\pagestyle{empty}

\begin{abstract}
We present a real-time dense geometric mapping algorithm for large-scale environments. Unlike existing methods which use pinhole cameras, our implementation is based on fisheye cameras \zh{whose large field of view benefits various computer vision applications for self-driving vehicles such as visual-inertial odometry, visual localization, and object detection.} Our algorithm runs on \zh{in-vehicle PCs} at approximately 15 Hz, enabling vision-only 3D scene perception for self-driving vehicles. For each synchronized set of images captured by multiple cameras, we first compute a depth map for a reference camera using plane-sweeping stereo. To maintain both accuracy and efficiency, while accounting for the fact that fisheye images have a lower angular resolution, we recover the depths using multiple image resolutions. We adopt the fast object detection framework, YOLOv3, to remove potentially dynamic objects. At the end of the pipeline, we fuse the fisheye depth images into the truncated signed distance function (TSDF) volume to obtain a 3D map. We evaluate our method on large-scale urban datasets, and results show that our method works well in complex dynamic environments.  
\end{abstract}

\section{Introduction}
Real-time 3D mapping of an environment is required for autonomous vehicles to perceive and thus navigate in complex environments. Typically, LiDAR sensors are used for 3D perception as they can generate accurate 3D point clouds in real-time. 
In contrast to LiDAR sensors, cameras do not directly provide 3D information. However, we can recover 3D information from multiple images based on multi-view geometry techniques. In comparison to LiDAR sensors which suffer from poor vertical resolution, cameras can generate 3D maps with high resolution. Moreover, scene understanding techniques for cameras are more well-developed than those for LiDAR point clouds \cite{ramanishka2018toward}. 

Image-based 3D dense mapping usually comprises two key steps. The first step involves recovering depth information via a (multi-view) stereo algorithm. 
The second step entails temporal fusion of individual depth images with associated camera poses into a 3D map representation. Both monocular and binocular systems are used in previous work. Monocular systems \citep{schops2017large, Haene2015IROS} recover the depth with a sequence of consecutive frames, and have difficulty in handling moving objects. Binocular systems \citep{Pillai2016ICRA, pire2018Robotica} compute the depth from two images captured at the same time, and better handle complex \fvzh{dynamic} environments. 
Recently, \citet{Barsan2018ICRA} proposed a robust dense mapping algorithm for large-scale dynamic environments. Leveraging both instance-level segmentation and scene flow estimation, the system reconstructs both the static part of the environment and moving objects. However, the entire pipeline is too computationally expensive to be applied to self-driving vehicles, especially at high driving speeds. \zh{Constrained by the instance-aware semantic segmentation phase, their method can only run at approximately 2.5Hz on stereo pinhole cameras.}

\begin{figure}[t]
	\centering
	\begin{tabular}{*{4}{c@{\hspace{2px}}}}
		\includegraphics[width=0.23\textwidth]{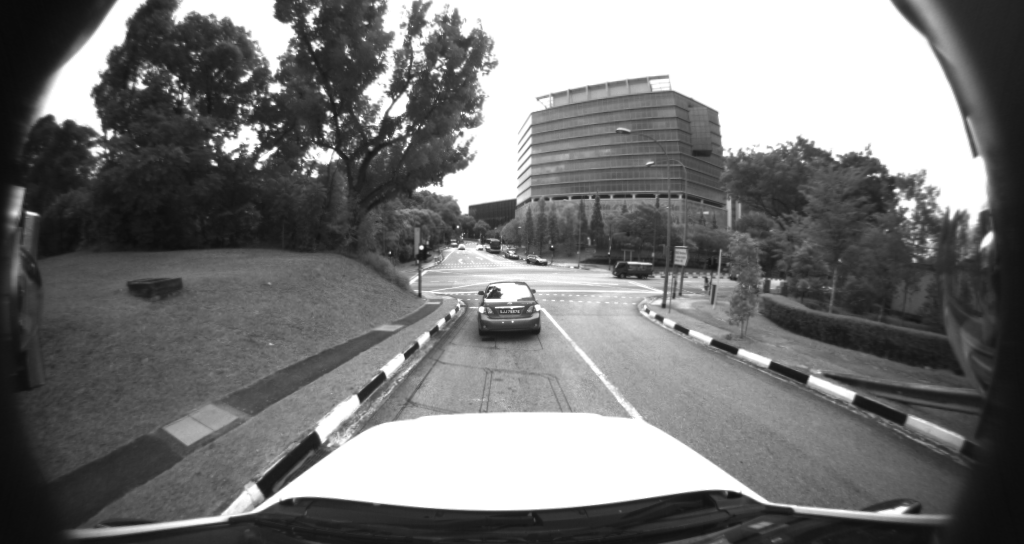} &
		\includegraphics[width=0.23\textwidth]{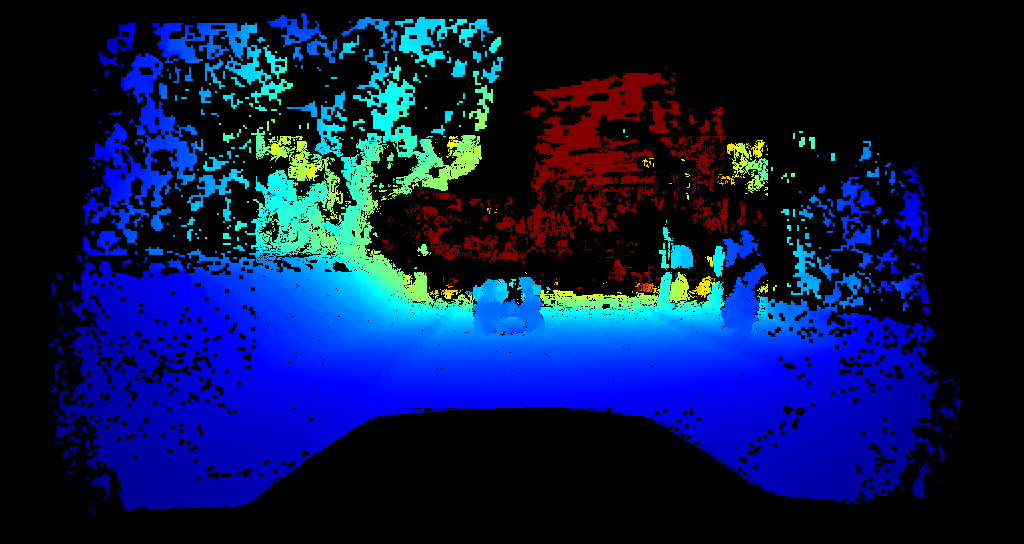} \vspace{-0.2em}\\
		\small{(a)} & \small{(b)} \\
		\includegraphics[width=0.23\textwidth]{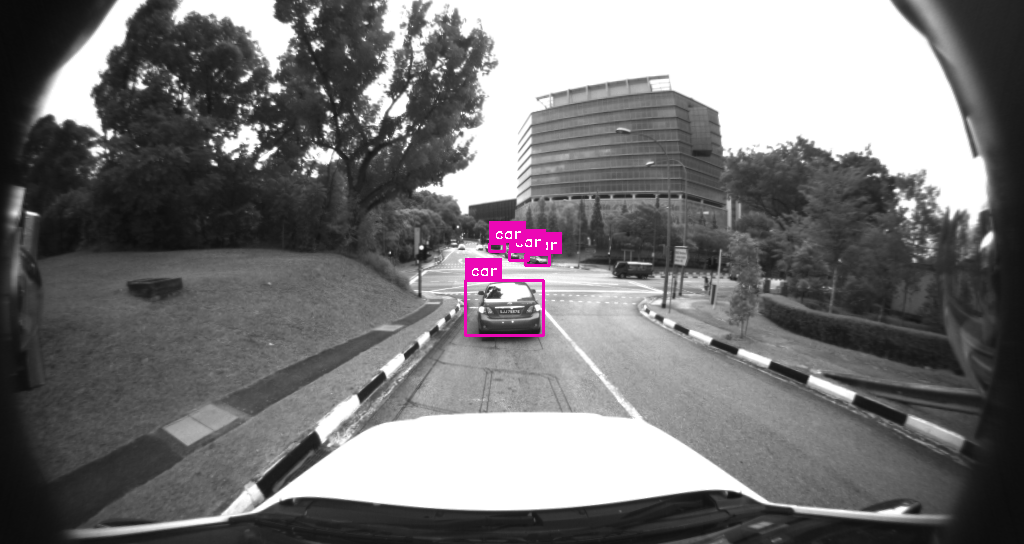} &
		\includegraphics[width=0.23\textwidth]{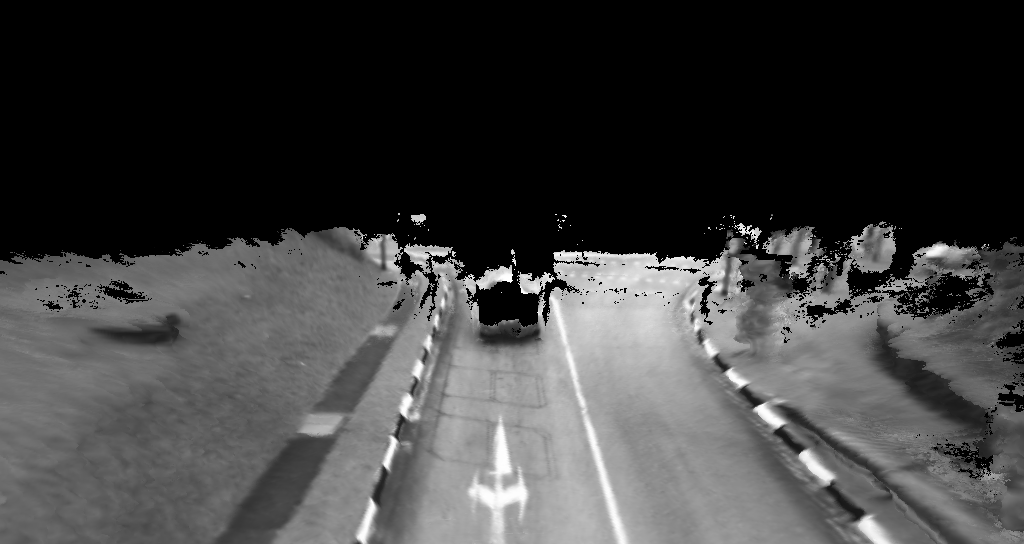} \vspace{-0.2em}\\
		\small{(c)} & \small{(d)}\\
		\vspace{-1.5em}
	\end{tabular}
	\caption{Input and outputs of our real-time dense mapping system for fisheye cameras. (a) Image from the front center camera; (b) Depth map generated by plane-sweep stereo; (c) Dynamic objects detected by the object detector; (d) The 3D map built by our approach, with dynamic objects automatically removed from the model.}
	\label{fig:teaser}
	\vspace{-1.5em}
\end{figure}

In this paper, we present a real-time dense mapping system for self-driving vehicles. 
\zh{Due to the large field of view, fisheye cameras have shown better performance than pinhole cameras for multiple tasks including visual-inertial odometry \citep{Liu2018IROS} and localization \citep{Geppert2019ICRA}. Hence, we adopt a multi-fisheye-camera stereo setup for dense mapping with a self-driving vehicle. However, fisheye cameras have a drawback: 
\fvzh{the pixels in the central region of a fisheye image have lower angular resolution compared to those of a pinhole image with the same image resolution \cite{zhang2016benefit}}. 
This lower resolution degrades the depth estimation of far-away objects in front of vehicles, and this degradation is made more severe with downsampling of fisheye images for real-time processing.}
In order to maintain both accuracy and efficiency, we propose a new strategy for fisheye depth map recovery using images with different resolutions. \zh{Several filtering methods are adopted to filter noisy and unreliable depth estimates in texture-poor and low-resolution areas.} Subsequently, we fuse the fisheye depth maps directly into a 3D map model. To increase the system's scalability, we reduce both memory usage and run-time with local map pruning and only store map data in the vehicle's vicinity. In contrast to \citet{Barsan2018ICRA}, in order to fulfill the requirement of real-time mapping for self-driving vehicles,  we adopt a fast object detection pipeline to handle potentially moving objects. We filter these objects directly from the depth map before the depth fusion in order to obtain a static map of the surrounding environment. Experiments show that although there are occasional false positive detections which filter out static parts of the environment, there is little impact on the 3D mapping as depth information is fused temporally. \zh{The on-road test shows that our method is capable of generating dense maps in real-time, and with good accuracy and reasonable completeness when the car drives up to 40 km/h.}

In summary, this paper makes the following contributions: 
(1) We propose a practical system to achieve real-time dense mapping for self-driving vehicles purely using fisheye cameras. (2) A new multi-scale strategy is proposed for fisheye depth map estimation to maintain both depth map accuracy and efficiency. (3) Multiple depth filtering and local map pruning techniques are \fvzh{studied and} evaluated with LiDAR data, giving us insights into directions for future work. 
\section{Related work}

Many works \citep{Geiger2011IV, Pillai2016ICRA, pire2018Robotica} exist for real-time dense mapping with multi-camera systems including stereo cameras. 
\fvzh{These works are based on the stereo matching using a pinhole camera model,  which assumes rectified input images and performs disparity search along the epipolar line. So they cannot be extended to fisheye cameras, and rectifying the fisheye images to pinhole images would lead to a significant loss of field-of-view \citep{Haene20153DV}. \citet{Pollefeys2008IJCV} propose a method of 3d reconstruction from video which assumes that the scene is static, while this will lead to dynamic objects leaving a trail of artifacts in the 3D map.}
\citet{HernandezJuarez2017BMVC} convert a depth map into a compact stixel representation; this representation only works well for urban environments with strong planarity features unlike a TSDF volume representation \citep{Kahler2015ISMAR, Newcombe2011ISMAR} which can represent arbitrary environments. \citet{schops2017large} implement large-scale 3D reconstruction with a monocular fisheye camera and TSDF-based depth fusion. This reconstruction pipeline is not able to handle dynamic environments too. \citet{Barsan2018ICRA} propose a mapping approach for dynamic environments but use stereo pinhole cameras, and their approach is too slow for real-time 3D perception for self-driving vehicles.
\fvzh{Recently, several visual odometry methods \cite{caruso2015large, heng2016semi, liu2017direct} for fisheye cameras have been proposed, and the camera poses recovered from these methods could be utilized for our depth map fusion. }

\section{Real-time Dense Mapping}

\begin{figure}[t]
	\centering
	\hspace{-0.2em}\includegraphics[width=0.49\textwidth]{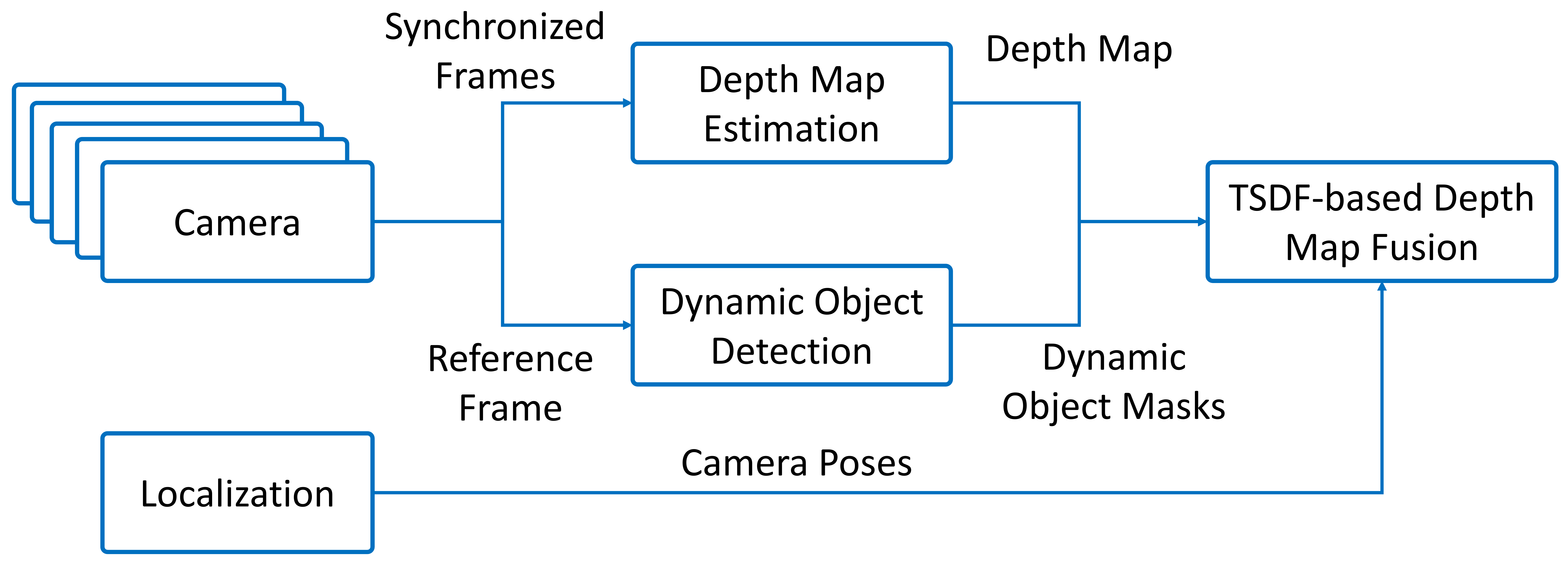}
	\vspace{-1.5em}
	\caption{Given multiple images captured at the same point in time by fisheye cameras, our approach first computes depth maps for a reference frames. Detecting dynamic objects allow us to avoid artifacts in the 3D model caused by moving objects. Using camera poses provided by a localization system, the resulting depth maps are integrated into a truncated signed distance function volume.}
	\label{fig:pipeline}
	\vspace{-1.5em}
\end{figure}

In this section, we describe our real-time dense mapping method for fisheye cameras. As shown in \figref{fig:pipeline}, the proposed method consists of three steps: depth map estimation, dynamic object detection, and TDSF-based depth map fusion. At each time step, we obtain synchronized frames from multiple fisheye cameras. 
We use stereo matching between the different camera images and a reference camera to estimate a depth map for the reference camera. At the same time, we input the image from the reference camera into an object detector and obtain the masks of all potentially moving objects. Then we use the detection results to mask out potentially moving objects from the depth map to avoid introducing artifacts into the 3D map. 
Finally, we fuse the resulting depth map into the TSDF volume to reconstruct a 3D map of the static environment. 

\subsection{Depth Map Estimation at Multiple Scales}
To enable large-field-of-view mapping which is advantageous for self-driving vehicles, we use multiple fisheye images for depth map computation. We use plane-sweeping stereo \cite{Collins1996CVPR} for the depth map estimation. Plane-sweeping stereo matches the reference image to a set of supporting images with known relative poses by sweeping a set of planes through 3D space. Each position of a plane defines a homography mapping pixels from the reference image into other images. At the same time, each plane position corresponds to a depth hypothesis for each pixel. For a given pixel in the reference image, the quality of a depth hypothesis can be evaluated by computing the image dissimilarity within a small local window. As shown in \citet{Haene20153DV}, the image warping function can be run on the fly on a graphics processing unit (GPU) at little additional computational costs. 
 Compared to standard stereo matching methods \cite{Geiger_ACCV_2010} and recent deep-learning-based methods \cite{Mayer2016CVPR}, it is not necessary for the plane-sweeping stereo algorithm to undistort the fisheye images prior to depth map computation which would otherwise result in a loss of field of view~\cite{Haene20153DV}. In addition, plane-sweeping stereo allows us to easily use more than two cameras. 

As proposed in \citet{Gallup2007CVPR} and \citet{Haene2015IROS}, we sweep planes along multiple directions. As is standard, one set of planes is parallel to the image plane of the reference camera. The other set of planes is parallel to the ground plane, which leads to a better estimate of the ground as observed in~\citet{Haene2015IROS}. The extrinsic calibration of the cameras provides an estimate for the ground plane. Subsequently, only a few planes close to the ground are used for this sweeping direction.

For our experiments, we use 64 fronto-parallel planes and 30 planes parallel to the ground. The negative zero-normalized cross-correlation (ZNCC) score over a local window is used as the matching cost to measure the image disimilarity. A matching cost of 1 corresponds to a ZNCC score of $-1$ while a matching cost of 0 corresponds to a ZNCC score of 1. We use a local window of $9 \times 9$ pixels for  full resolution and a window of $7 \times 7$ pixels for downsampled images.


\PAR{Multi-scale strategy:} 
Depth estimation is a time-consuming task even on a GPU. 
Thus, downsampling the fisheye images to a lower resolution is necessary for achieving real-time performance. 
Compared to pinhole images, the central portion of fisheye images have a lower angular resolution. 
Together with downsampling of fisheye images, it is challenging to accurately reconstruct objects far from the scene, which in autonomous driving scenarios are typically in the center of view of forward facing cameras. 
At the same time, obtaining accurate depth estimates for objects in the center of view is important for path planning and collision detection in the context of self-driving cars. 
In order to both reconstruct far-away objects and cover the full field-of-view of the fisheye images, 
 we propose a new multi-scale strategy for fisheye depth image estimation:  
First, we run depth estimation on downsampled fisheye images, and subsequently, up-sample the depth estimates to the original resolution. Next, we crop the center area of the fisheye image at the original resolution and compute the depth image corresponding to the cropped area. We then fuse the two depth maps. 
Intuitively, our strategy corresponds to combining a low-resolution depth map generated from fisheye images with a higher-resolution depth map obtained by close-to-pinhole cameras (as the center of a fisheye image typically exhibits low radial distortion). 
\zh{In our experiments, the full image resolution is $1024 \times 544$ pixels. We downsample images to $512 \times 272$ pixels to obtain low-resolution images, and the cropping size is $ 572 \times 332$ pixels.}

\PAR{Depth map filtering:} 
As detailed in \secref{sec:fusion}, depth maps estimated at each point in time are fused into a single volume to obtain a 3D map of the scene. 
The raw depth maps generated by plane-sweeping stereo often contain outliers. 
Integrating outlier measurements into the volume leads to false predictions for the scene geometry, which in turn can cause problems for the path planning module which depends on accurate 3D maps. 
Thus, we perform several filtering steps designed to identify and remove outlier measurements. 

\begin{figure*}[t]
	\centering
	\begin{tabular}{*{4}{c@{\hspace{2px}}}}
		\includegraphics[width=0.24\textwidth]{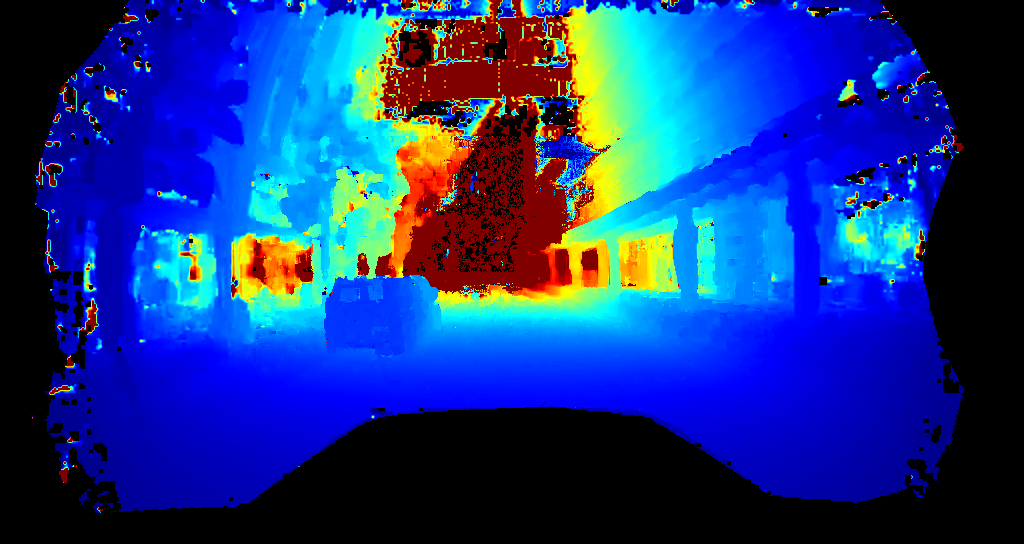} &
		\includegraphics[width=0.24\textwidth]{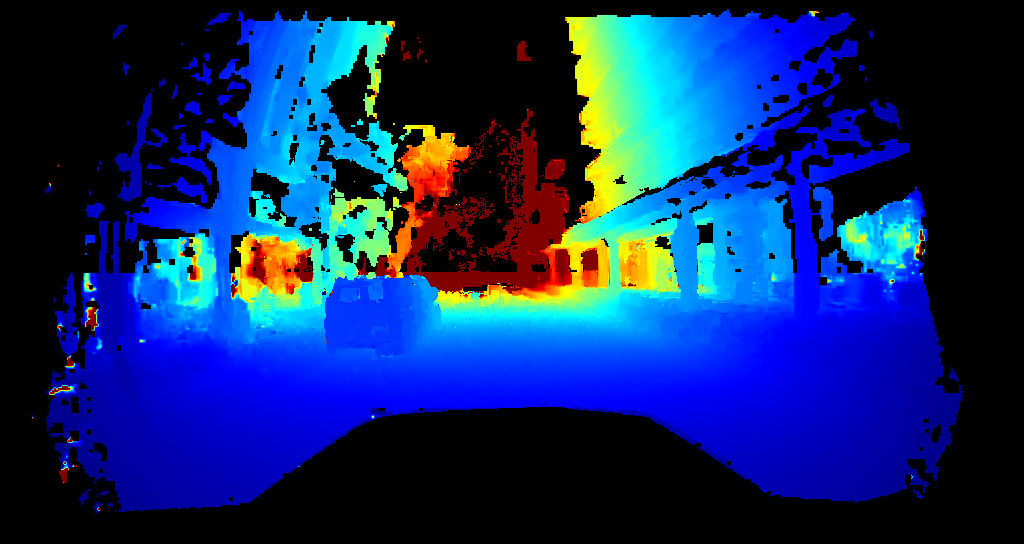} &
		\includegraphics[width=0.24\textwidth]{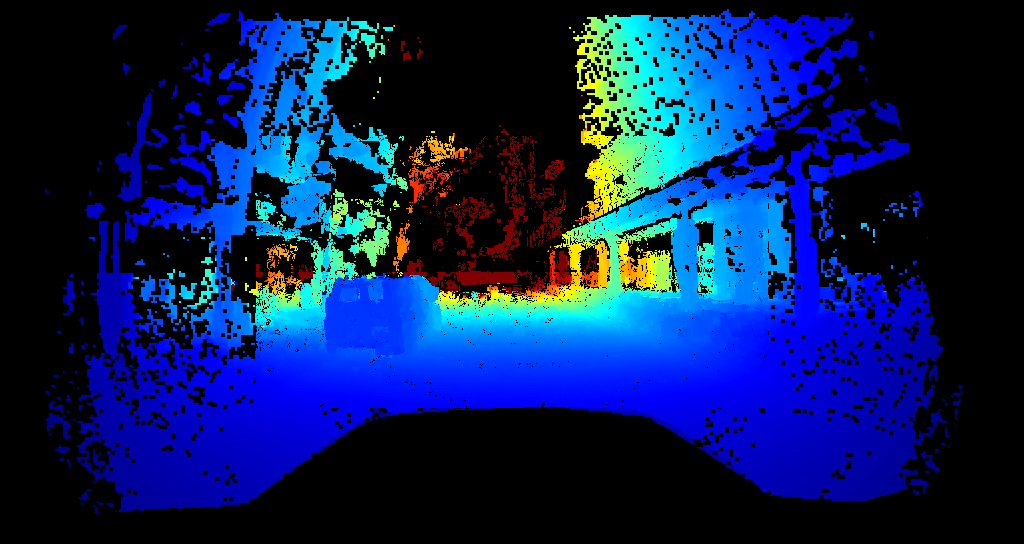} &
		\includegraphics[width=0.24\textwidth]{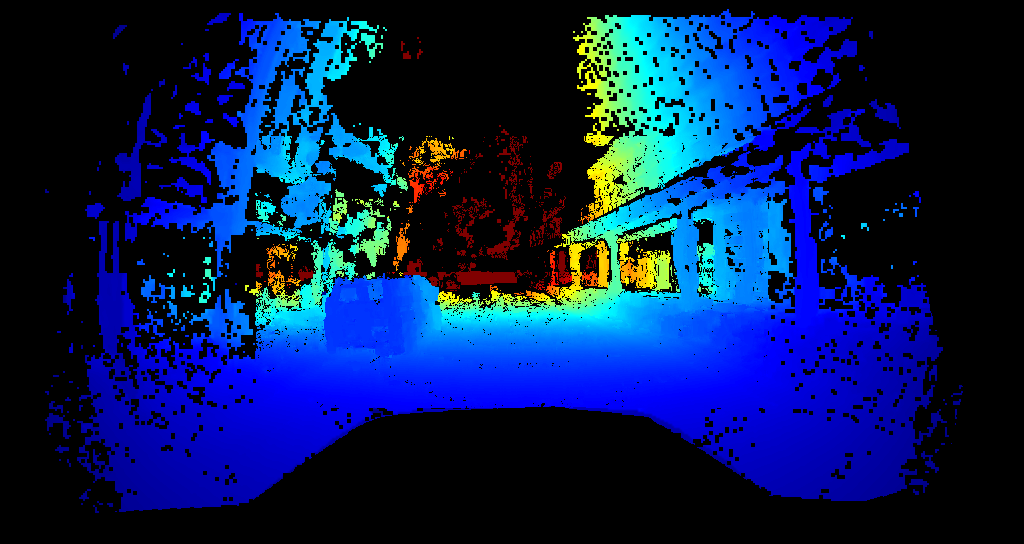} \vspace{-0.2em}\\
	   
		\small{(a)} & \small{(b)} & \small{(c)} & \small{(d)}\\
		\vspace{-1.8em}
	\end{tabular}
	\caption{Illustrating the impact of depth map filtering: (a) Raw depth image; (b) Depth image after filtering based on best matching cost; (c) Depth image after filtering based on the ratio between the first and second best costs; (d) Depth image after local continuity checking.}
	\label{fig:effect_filtering}
	\vspace{-1.8em}
\end{figure*}

Firstly, we filter the depth using the matching cost value of the best depth candidate for a pixel. If the cost is larger than $\alpha$, we consider it an outlier. Then, the ratio between the first and second best cost values is used to further remove potentially unreliable depths. The larger the ratio is, the more unique and reliable the estimated depth is. Finally, we use local depth continuity checking to filter out noisy depth values. For each pixel, we compare its depth value against the depths of its neighboring pixels in a local window. If the difference is smaller than a threshold $\gamma$, we will consider the neighboring pixels to be consistent with the central pixel. If the ratio of the consistent pixels in the local window is smaller than $\delta$, we will consider the center pixel to have inaccurate depth. \zh{In our implementation, we set $\alpha$ for the upper and lower parts of the image as 0.05 and 0.3 respectively, and set $\gamma$ and $\delta$ to be $0.5m$ and 0.3.} The effect of these filtering steps is shown in \figref{fig:effect_filtering}. From this figure, we can see that most of the unreliable depth information \fvzh{(e.g., in the sky, featureless building facades and long distance areas)} is filtered out successfully. 

\subsection{Dynamic Object Detection}
In order to deal with moving objects, \citet{Barsan2018ICRA} used an instance-level semantic segmentation method \cite{dai2016instance}. 
Unfortunately, the method is too slow to achieve real-time performance even on a powerful GPU, let alone the GPU used on our vehicle. 
The reason is that it, as other instance-level segmentation methods, is based on a two-stage object detection approach which first generates many bounding box proposals before estimating a segmentation mask for each proposal. As a result, it is difficult for these methods to run at high speeds. Single-stage detectors are significantly faster than two-stage detectors as they treat object detection as a simple regression problem by learning the class probabilities and bounding box coordinates directly from the input image. Thus, we use a  single-stage \fvzh{object detection} method \citep{redmon2018yolov3} to detect potentially moving objects. Although we do not obtain as accurate object masks, we find that the single-stage method is sufficient in practice to filter out moving objects.

YOLOv3 \citep{redmon2018yolov3}  takes an image and divides it into a $S\times S$ grid. In each grid cell, it generates $B$ bounding boxes. For each bounding box, the network outputs a class probability and bounding box attributes. Finally, bounding boxes with associated class probabilities above a threshold value are chosen.

The datasets \citep{Lin2014ECCV} typically used to train object detectors are composed of pinhole color images. 
In contrast, we use grayscale fisheye images as the input to our algorithms. 
In order to avoid having to create a large dataset with object annotations, we 
base our model on an existing one trained on the Microsoft COCO dataset \citep{Lin2014ECCV}. 
We adapt it to our dataset by truncating the first and last layers and fine-tune it on a small annotated dataset of images captured from our vehicle. 

We \zh{consider all kinds of vehicles \citep{Cordts2016Cityscapes} as potentially moving objects. We} labeled 2609 fisheye images and approximately 11200 different vehicles. The network is implemented in TensorFlow \citep{abadi2016tensorflow} and optimized using the Adam optimizer \citep{kingma2014adam}. We take two training steps. At first, we only train the first and last layer with a learning rate of  $10 e^{-4}$ for 10 epochs, and then fine-tune all the layers with a learning rate of $10 e^{-5}$ for 20 epochs.


\subsection{TSDF-based Depth Map Fusion}
\label{sec:fusion}
The depth maps only maintain local geometric information. 
To obtain a global 3D map, we need to fuse depth maps captured at different points in time (and thus from different camera poses) into a single 3D map. We use a standard fusion technique. \zh{The camera poses can be obtained either from visual-inertial odometry (VIO) or a GNSS/INS system.}
The scene is represented via a set of voxels, where each voxel stores a truncated signed distance function (TSDF) value \citep{Curless1996SIGGRAPH}. 
Thus, each voxel stores the signed distance to the closest object surface  (negative inside of objects, positive outside of objects, zero on surfaces), truncated to a certain maximum / minimum distance. 
We use the map fusion pipeline from the InfiniTAM library \citep{Kahler2015ISMAR,Kahler2016RAL} which consists of four stages. At first, new voxel blocks are allocated based on the current depth map. Secondly, the list of voxel blocks that are currently visible is updated. Subsequently, the new 3D information is fused into each of the visible voxel blocks. At last, voxel blocks can be chosen to be swapped out from the GPU to preserve memory.  
This last stage is important when mapping a large area. 

In contrast to \citet{Kahler2015ISMAR,Kahler2016RAL}, we take fisheye depth images as inputs instead of pinhole depth images. As in the plane-sweeping stereo stage~\cite{Haene20153DV}, the unified projective model for fisheye cameras \citep{mei2007single} is used during the voxel block allocation and integration to model our fisheye cameras. More specifically, when a new depth map arrives, we iterate through each pixel $\mathbf{p}$ with a valid depth value $d$ in the fisheye depth image, and compute its back-projected ray direction as
\begin{equation}
r(\mathbf{p}) = \left[
\begin{tabular}{c}
$\frac{\xi + \sqrt{1+ \left( 1- \xi^2 \right)(x^2+y^2)}}{x^2+y^2+1}x$  \\
$\frac{\xi + \sqrt{1+ \left( 1- \xi^2 \right)(x^2+y^2)}}{x^2+y^2+1}y$  \\
$\frac{\xi + \sqrt{1+ \left( 1- \xi^2 \right)(x^2+y^2)}}{x^2+y^2+1} - \xi$ 
\end{tabular}
\right] \enspace ,
\label{equ:ray_casting}
\end{equation}
where $(x, y)$ is the normalized coordinates of  $\mathbf{p}$, and $\xi$ is the mirror parameter of the fisheye camera \citep{mei2007single}. 
We consider the range 
from $d -\mu$ to $d +\mu$ along this ray for TSDF integration. 
We compute the coordinates of voxel blocks along this line segment and check whether these blocks have been allocated. If necessary, we allocate new voxel blocks. After updating the list of the currently visible voxel blocks, we do depth integration. 
To this end, each voxel $\mathbf{X}$ maintains a running average of the TSDF value $D(\mathbf{X})$. Given the current camera pose $(\mathbf{R}_d, \mathbf{t}_d)$, we get the voxel location in the camera coordinate system as $\mathbf{X}_d = \mathbf{R}_d\mathbf{X}+\mathbf{t}_d$. Then, we project $\mathbf{X}_d$ into the fisheye depth image using the unified projective model \citep{mei2007single}, and get its associated depth information. If the difference $\eta$ between the new depth value and 
the  associated depth value
is not smaller than $-\mu$, the SDF value will be updated as
\begin{equation}
\vspace{-0.5em}
D(\mathbf{X}) \leftarrow \left( w(\mathbf{X})D(\mathbf{X}) + \textup{min}\left(1, \frac{\eta}{\mu}\right) \right)/(w(\mathbf{X})+1),
\label{equ:updating}
\end{equation}
where $w(\mathbf{X})$ is a variable counting the number of previous observations in the running average, which will be updated and capped to a fixed maximum value.

In our system, we maintain a local map with a size of $60m \times 60m \times 3m$ centered at the current vehicle position for online mapping. 
We restrict ourselves to this area as this local 3D map is sufficient enough for path planning and obstacle detection and avoidance. 
In order to do this, we only allocate new voxel blocks within this region and swap old blocks that are no longer in this region. 
This reduces the memory consumption and allows our method to operate over a long period of time (as the map size is independent of the trajectory length driven by the vehicle). 
As a side benefit, it also helps to filter out a number of noisy observations as we observe that the depth estimation of the objects at long distances tends to be unreliable. 
To handle noisy depth information from plane-sweeping stereo, we only consider voxel blocks with at least 3 observations. 
For visualization, we use the fast raycasting algorithm from \citet{Kahler2015ISMAR,Kahler2016RAL}. 


\begin{figure}[t]
	\centering
	\begin{tabular}{*{2}{c@{\hspace{0px}}}}
		\hspace{-10px} \includegraphics[width=0.25\textwidth]{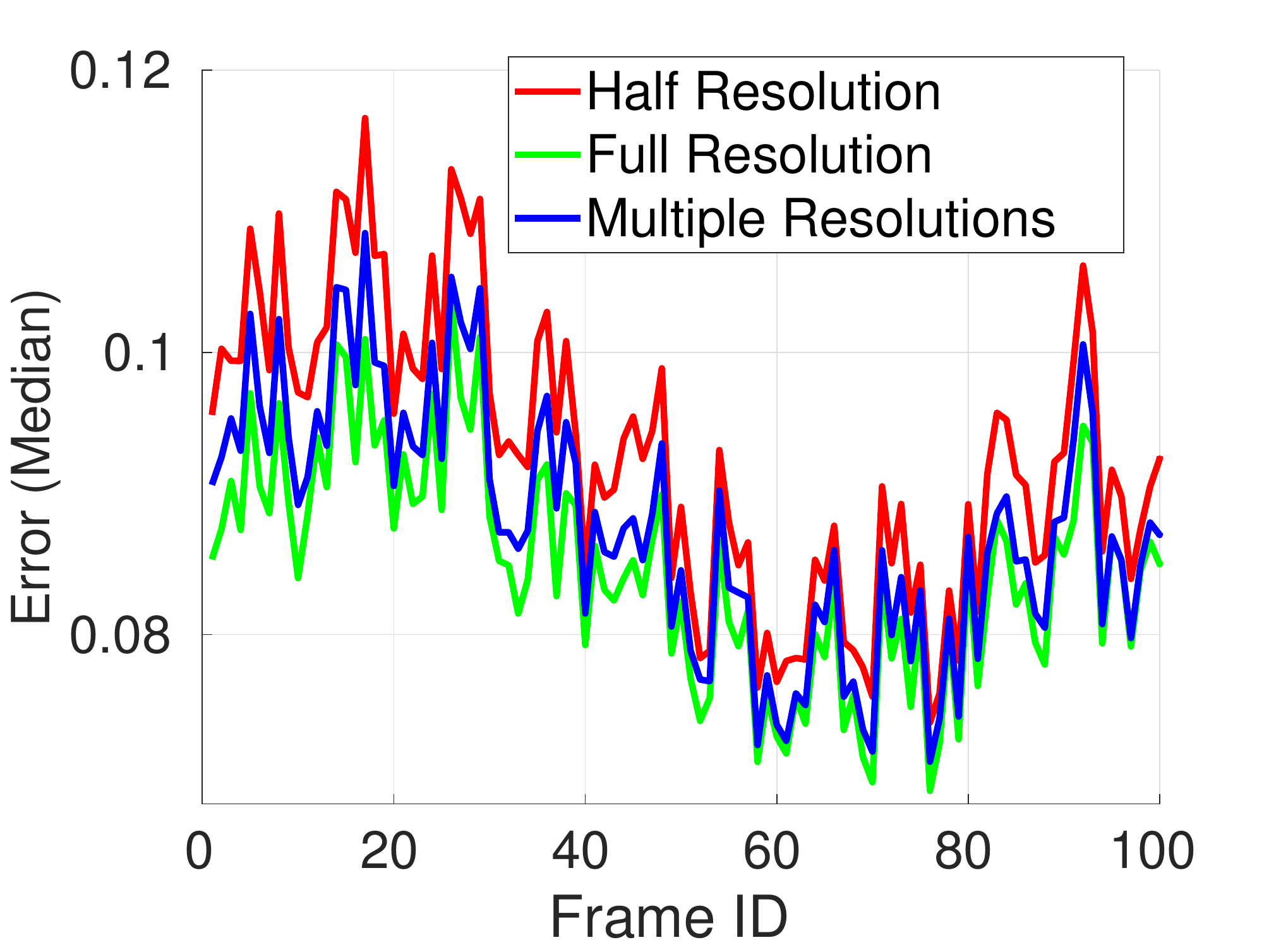} & \hspace{-5px}
		\includegraphics[width=0.25\textwidth]{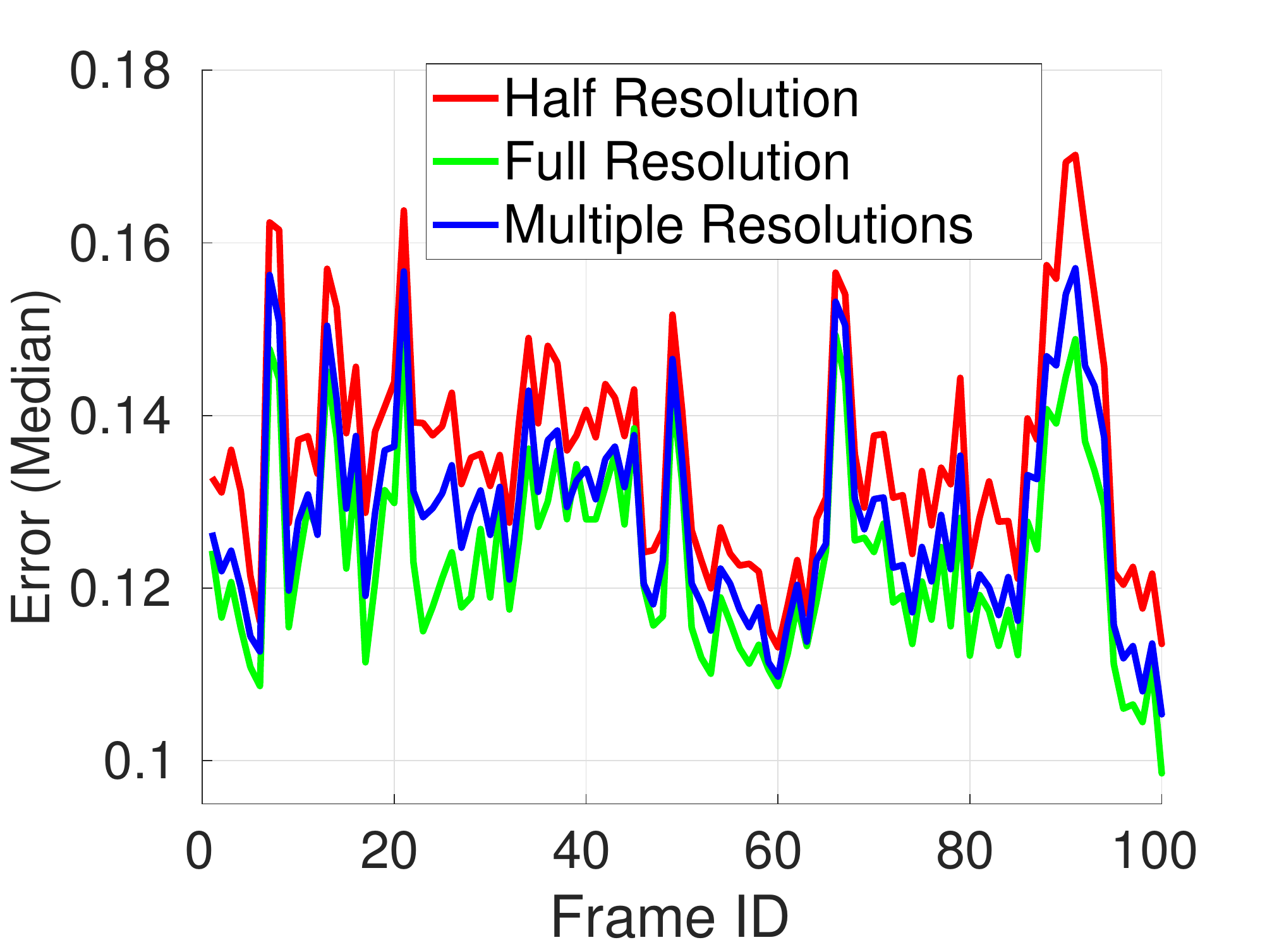} \vspace{-0.3em}\\
		\vspace{-0.5em}
		\small{(a) \em{South Buona Vista}} & \small{(b) \em{One North}}\\
	\end{tabular}
	\caption{The median value of the absolute errors (in meter) for each frame using different image resolutions as input to plane-sweep stereo. \fvzh{Our multi-scale strategy (``Multiple Resolutions") performs similar to using the original images (``Full Resolution"), and is better than ``Half Resolution".}}
	\vspace{-1.5em}
	\label{fig:medianErrResolution}
\end{figure}

\begin{figure*}[t]
	\centering
	\begin{tabular}{*{4}{c@{\hspace{6px}}}}
		\includegraphics[width=0.29\textwidth]{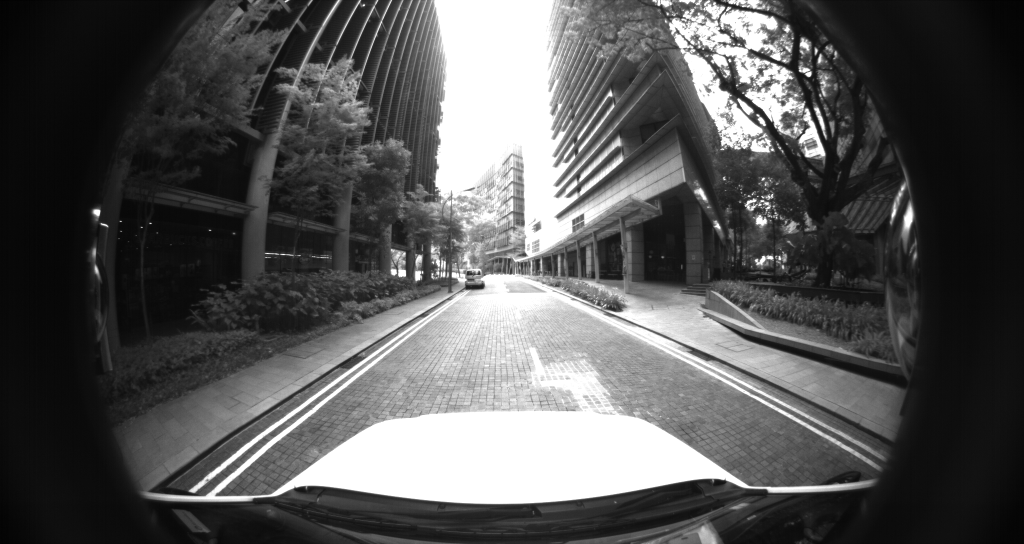} &
		\includegraphics[width=0.29\textwidth]{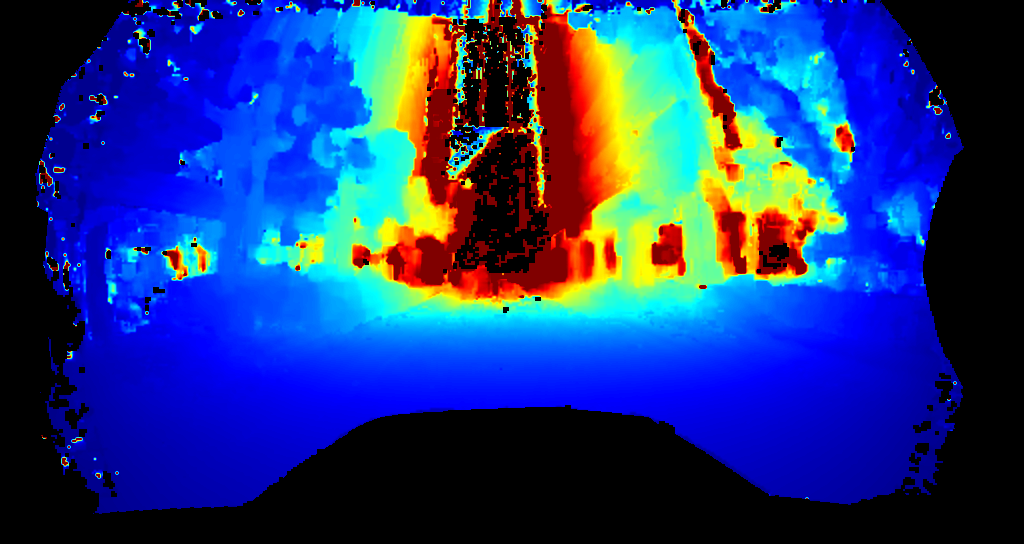} &
		\includegraphics[width=0.29\textwidth]{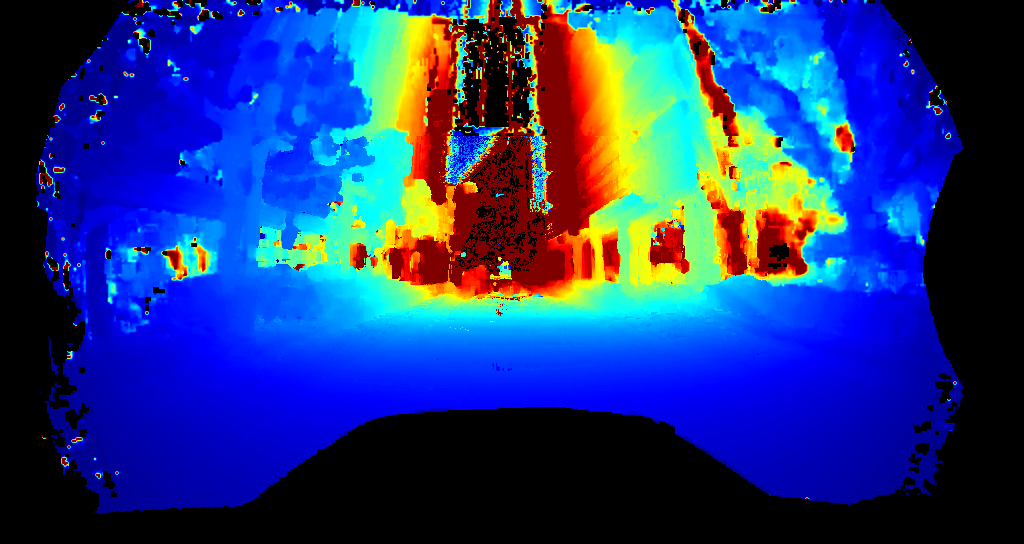} & \includegraphics[width=0.03\textwidth]{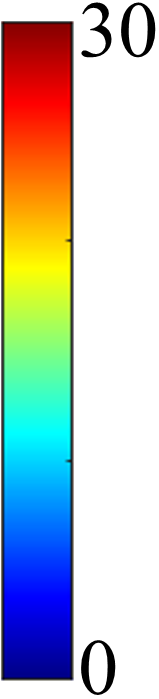} \vspace{-0.2em}\\
		
		\small{(a) \em{Reference image}} & \small{(b) Depth map from half resolution} & \small{(c) Depth map from multi-scale strategy} &\\
		\includegraphics[width=0.29\textwidth]{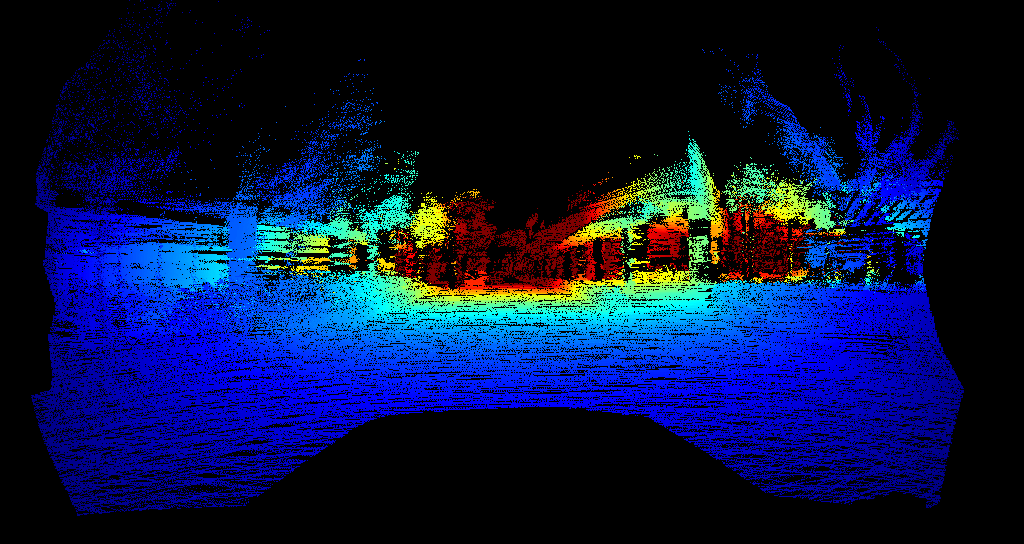} &
		\includegraphics[width=0.29\textwidth]{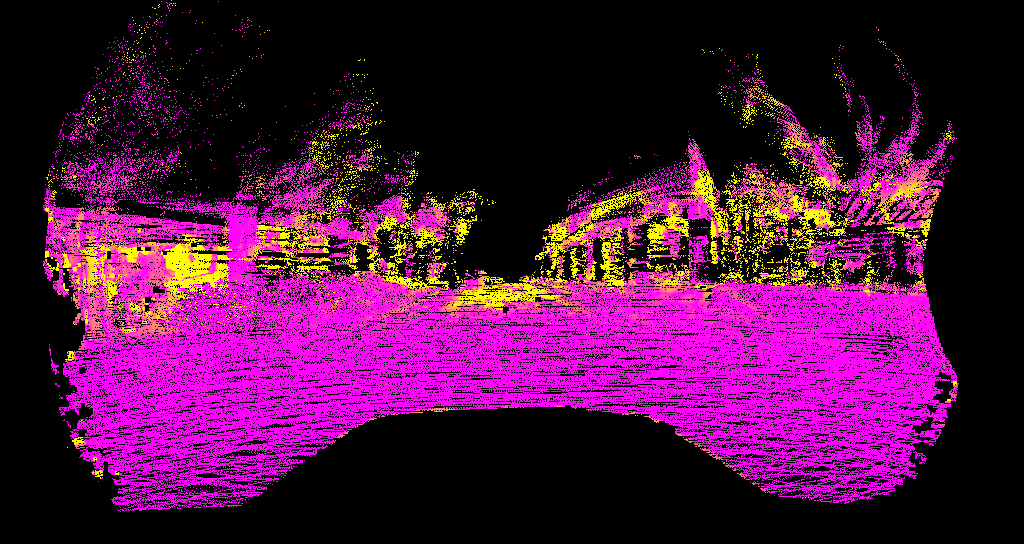} &
		\includegraphics[width=0.29\textwidth]{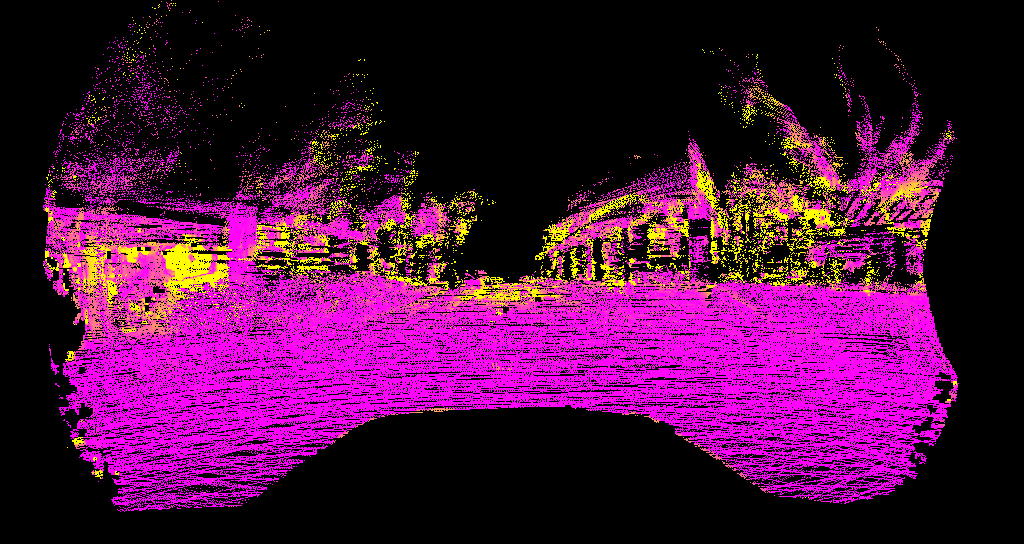}  & \includegraphics[width=0.03\textwidth]{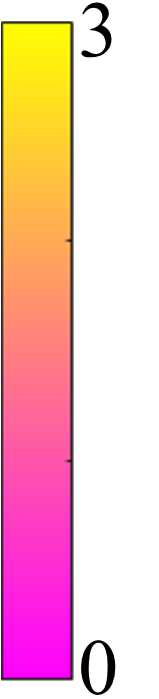} \vspace{-0.2em}\\
		\small{(e) Ground truth depth map} & \small{(f) Error map for half resolution}  & \small{(g) Error map for multi-scale strategy} &\\
		\vspace{-1.5em}
	\end{tabular}
	\caption{Visualizing depth maps and their errors for one frame from the  \emph{One North} sequence. \zh{The jet colormap is used for the depth map visualization with a range of $[0, 30m]$, and the spring colormap is used for the error map visualization with a range of $[0, 3m]$.} }
	\vspace{-1.8em}
	\label{fig:depthErrOne North}
\end{figure*}
\section{Experimental Evaluation}

\subsection{Experimental Setup}
We evaluate our real-time dense mapping pipeline on the self-driving testing platform
described in \citep{Heng2019ICRA}. Specifically speaking, we use five near-infrared (NIR) cameras mounted on top of the car and pointing to the front of the vehicle. These cameras output $1024\times544$ gray scale images at 25 frames per second, and are hardware-time-synchronized with the GNSS/INS system. 
There are also 3D LiDAR sensors mounted on top of the vehicle, and the fused point cloud data from these LiDAR sensors is used to evaluate the performance of our method. \zh{The camera poses computed from the GNSS/INS system are used in our experiments.}


In the vehicle, there are two PCs, each of which comes with an Intel i7 2.4 GHz CPU and a NVIDIA GTX 1080 GPU. In addition, we use a laptop with an Intel i7 2.8 GHz CPU and a NVIDIA Quadro M5000M GPU. The depth map estimation and dynamic object detection are each run on one PC, and the TSDF-based depth fusion is run on the laptop. The average running times per frame for these three components are about 60 ms, 40ms, and 20ms. As they are running in parallel, the frame rate of the whole pipeline is 15 Hz which is higher than most of the current LiDAR sensors. The current bottleneck is the depth map estimation stage. 
Parameter tuning, e.g., using fewer sweeping planes or smaller image resolutions, can potentially accelerate this stage, potentially at the cost of the depth map quality. Yet, our experience shows that our current frame rate is high enough for real-time mapping with a speed of up to 40km/h.

Our experiments are conducted in two different environments: \emph{South Buona Vista} and \emph{One North}. The \emph{South Buona Vista} sequence mainly follows a road passing through a forest and has a length of 3km. 
The \emph{One North} sequence follows roads in an urban area and has a length of 5km. 


\subsection{Evaluation of the Depth Estimation Stage}
\label{sec:evaluation_depth}
In order to qualitatively evaluate our depth maps, we fuse the 3D point cloud data from 2 LiDARs over a certain distance. We then project these points into the reference fisheye view to obtain a depth image which we use as ground truth. As moving objects adversely impact the ground truth depth image built from accumulation over time, we manually selected one sequence from each environment without moving objects for the evaluation. \fvzh{We compute the absolute difference between the estimated depth and the ground-truth depth, and calculate the mean and median values of the absolute differences in each frame for comparison.}

In the first experiment, we evaluate the performance of our multi-scale strategy \zh{without filtering steps}. We compute the depth maps using full-resolution and downsampled images, and using our multi-scale approach. 
\fvzh{\figref{fig:medianErrResolution} shows the median value of the absolute error of the depth maps} 
and  \figref{fig:depthErrOne North} shows the error map for a selected frame. 
As can be seen from \figref{fig:medianErrResolution}, using our proposed multi-scale strategy produces significantly better depth maps compared to simply using low-resolution images. 
The improvement mainly happens in image areas with large depth values as shown in \figref{fig:depthErrOne North}. 
Such image areas are mostly located in the center of the image. 
By using the full image resolution only in this image area, our approach nearly achieves the same quality as when using the full resolution for the complete image. 
\zh{The averaging running times of depth estimation without filtering for full resolution, down-sampled resolution and multi-scale approach are 50ms, 16ms and 36ms respectively, and our multi-scale approach reduces the running time by about $28\%$ compared to processing original full-resolution images.}


\begin{figure}[t]
	\centering
	\begin{tabular}{*{2}{c@{\hspace{0px}}}}
		\hspace{-10px} \includegraphics[width=0.25\textwidth]{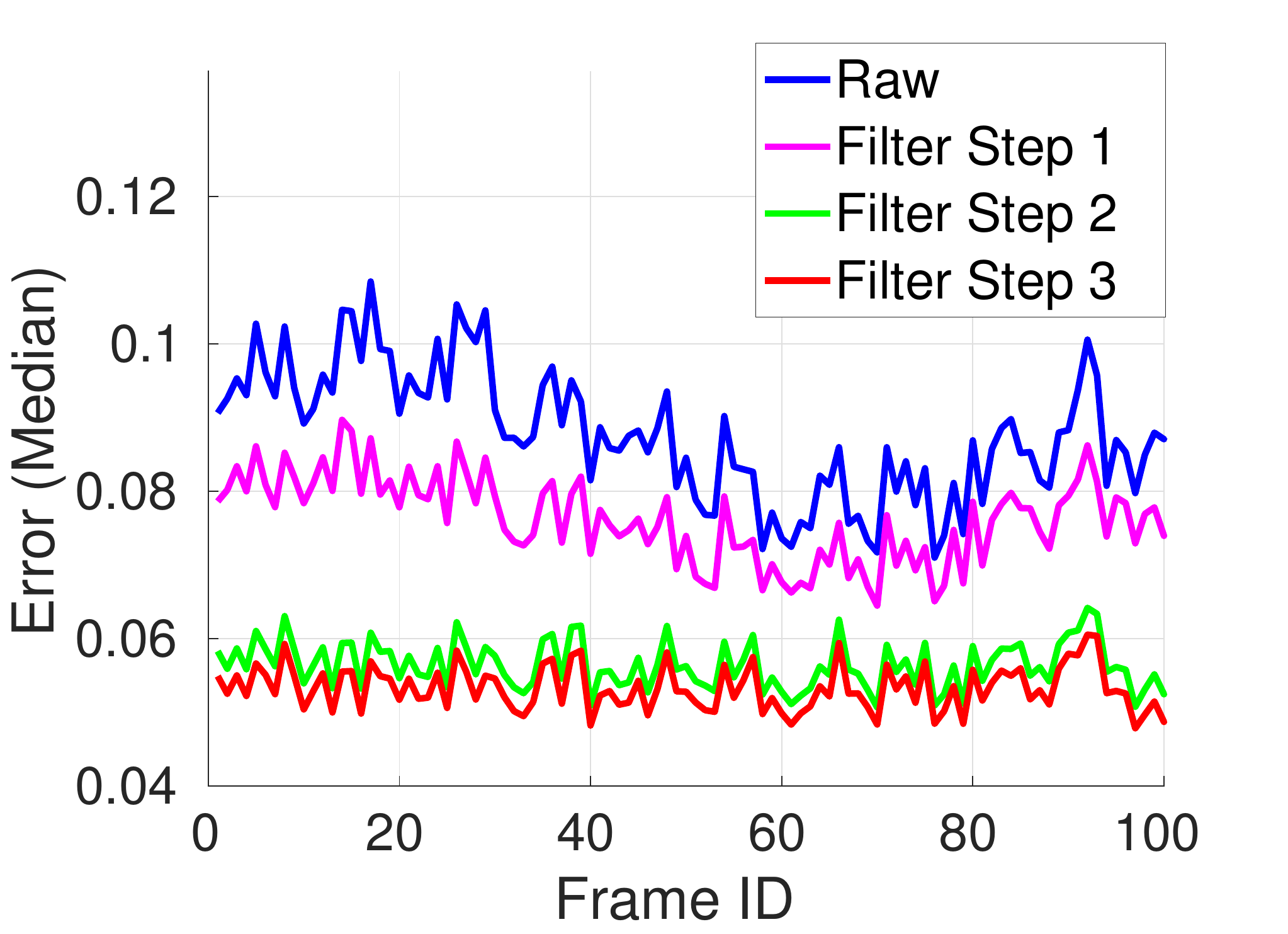} &
		\hspace{-10px} \includegraphics[width=0.25\textwidth]{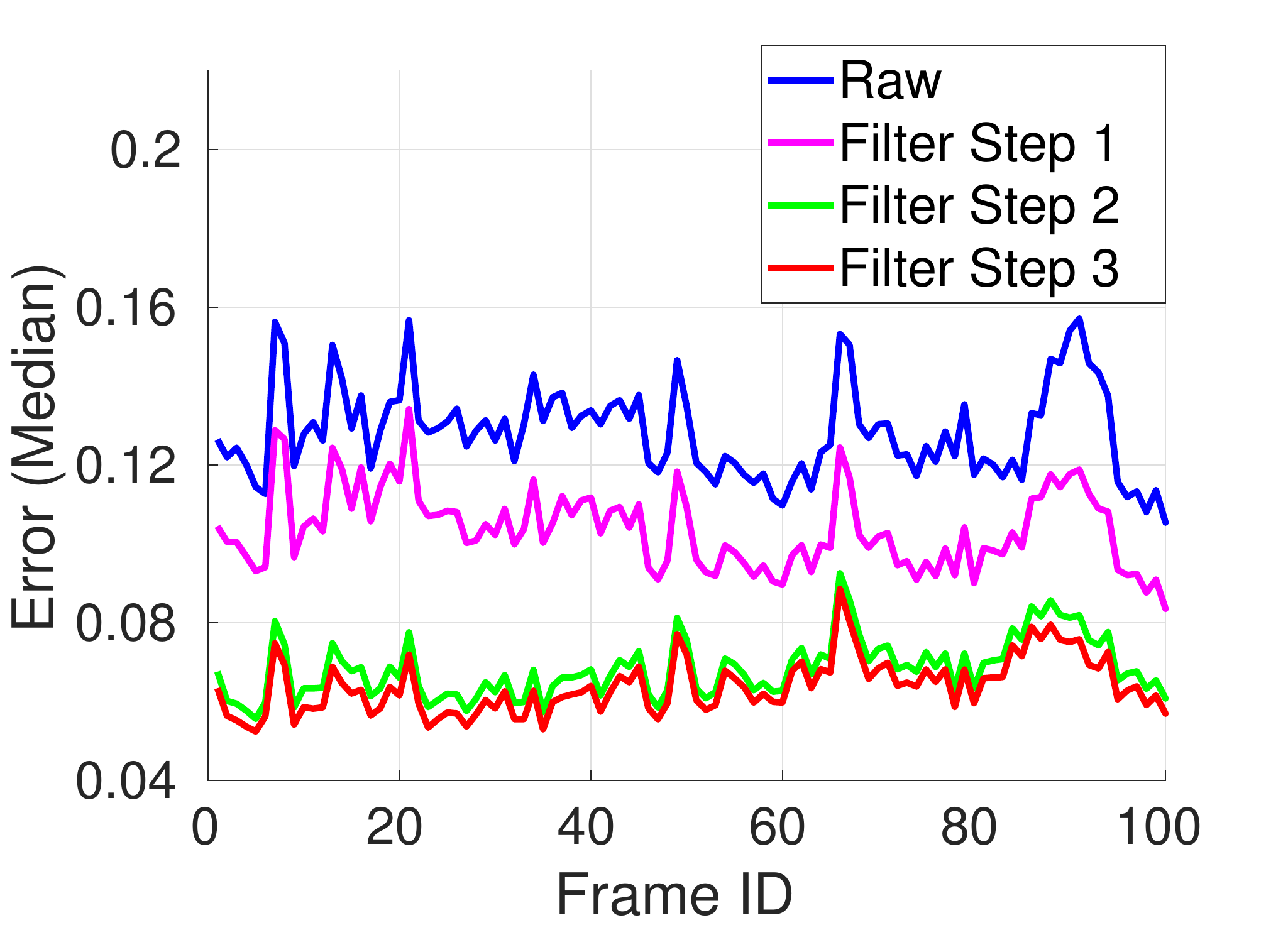} \vspace{-0.3em}\\
		
		\small{(a) \em{South Buona Vista}} \small{(Median)} & \small{(b) \em{One North}} \small{(Median)} \\
		\hspace{-10px} \includegraphics[width=0.25\textwidth]{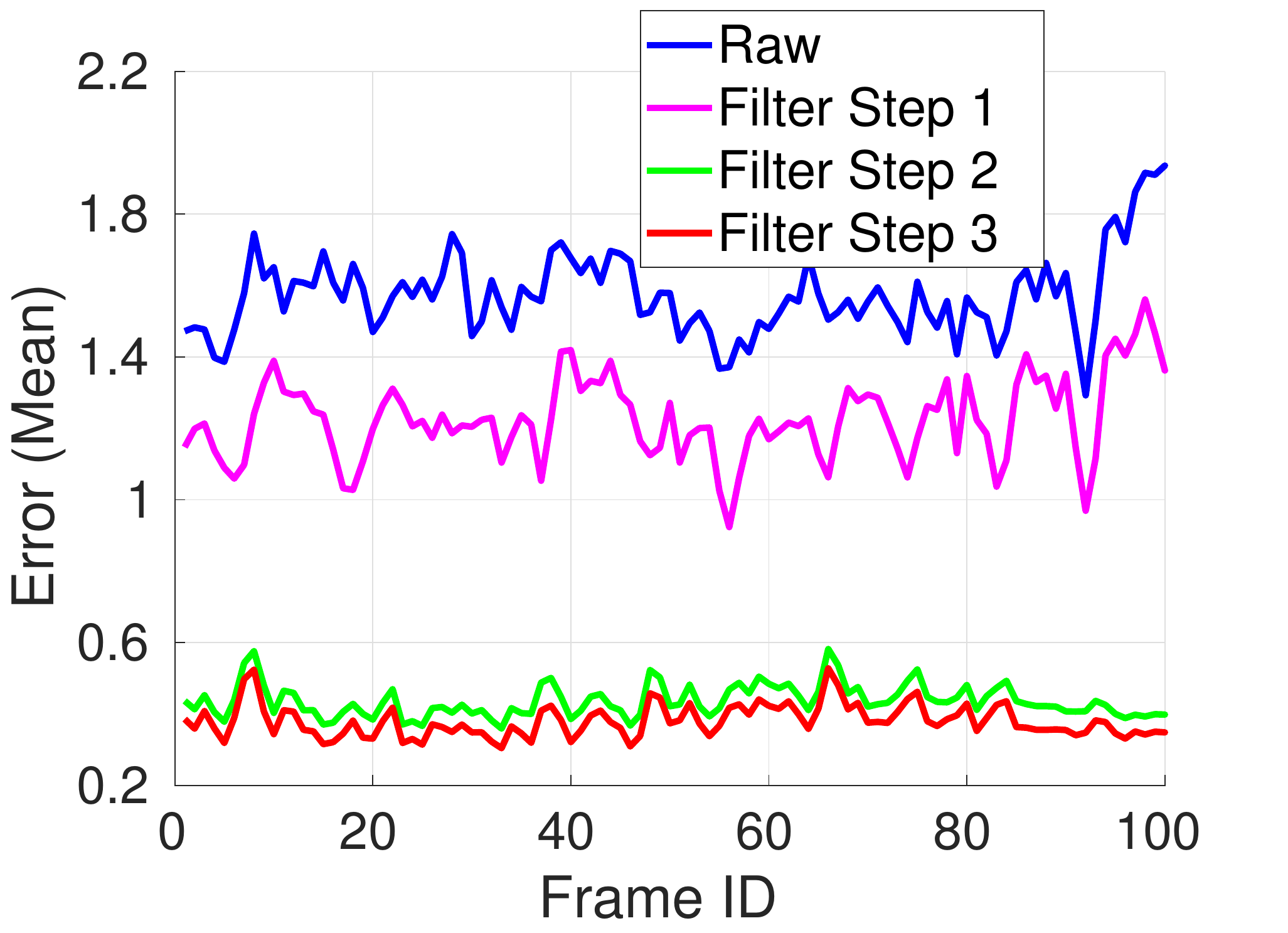} &
		\hspace{-10px} \includegraphics[width=0.25\textwidth]{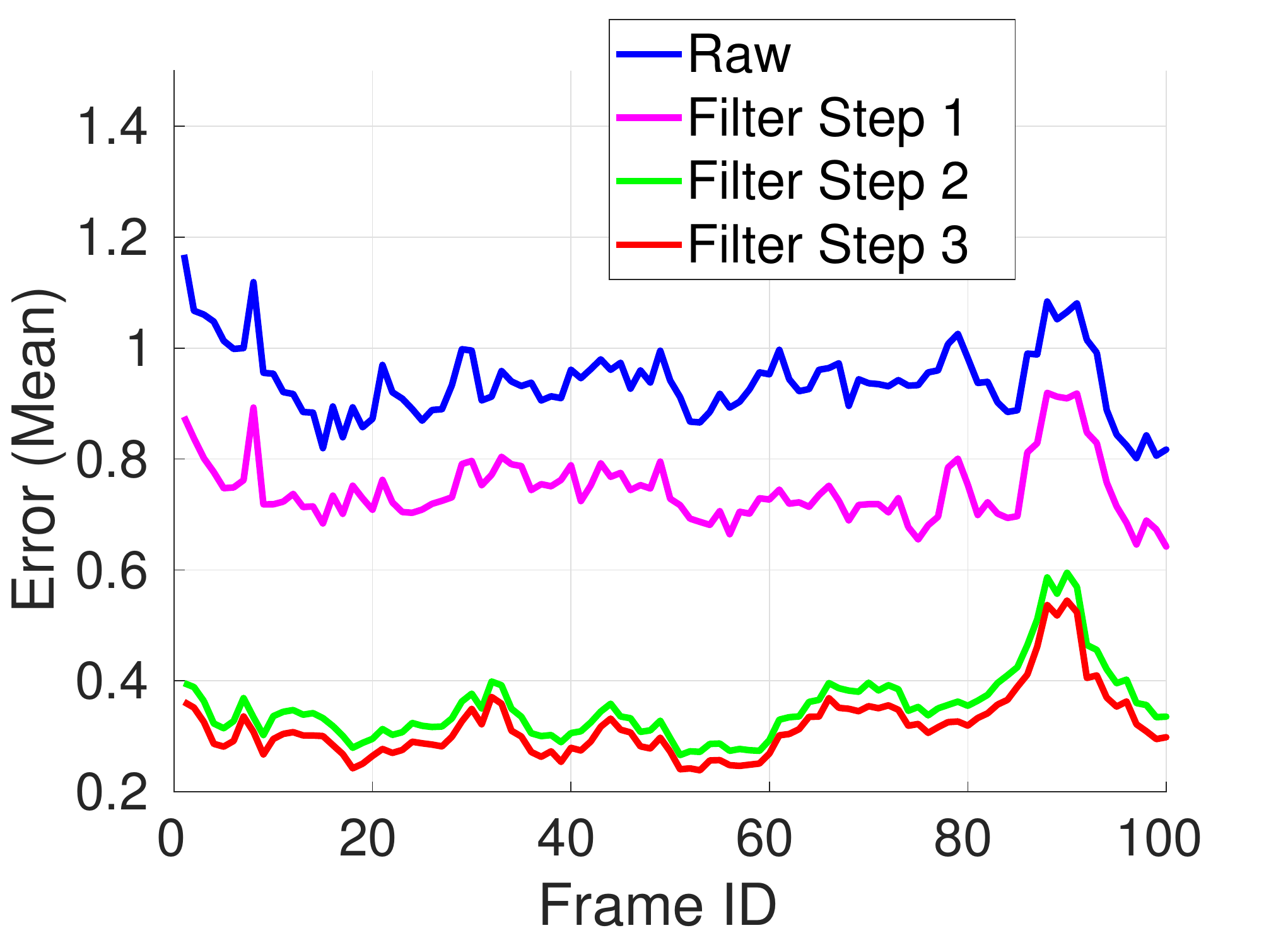} \vspace{-0.3em}\\
		\vspace{-0.5em}
		\small{(c) \em{South Buona Vista}} \small{(Mean)} & \small{(d) \em{One North}} \small{(Mean)} \\
	\end{tabular}
	\caption{Impact of the filtering stages on the depth error (in meter).}
	\label{fig:medianErrFilters}
	\vspace{-1.5em}
\end{figure}


\begin{figure}[t]
	\centering
	\begin{tabular}{*{2}{c@{\hspace{2px}}}}
		\includegraphics[width=0.24\textwidth]{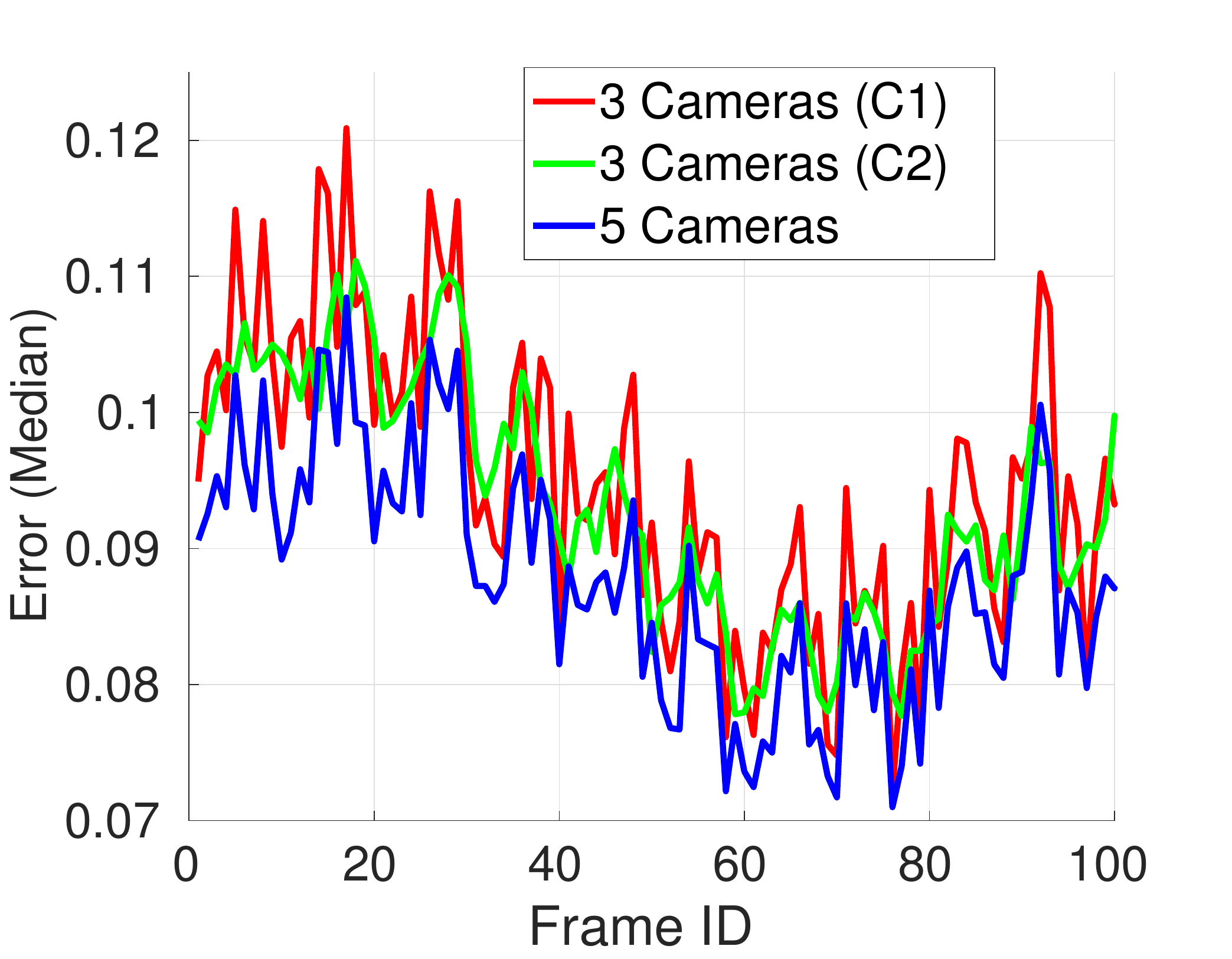} &
		\includegraphics[width=0.24\textwidth]{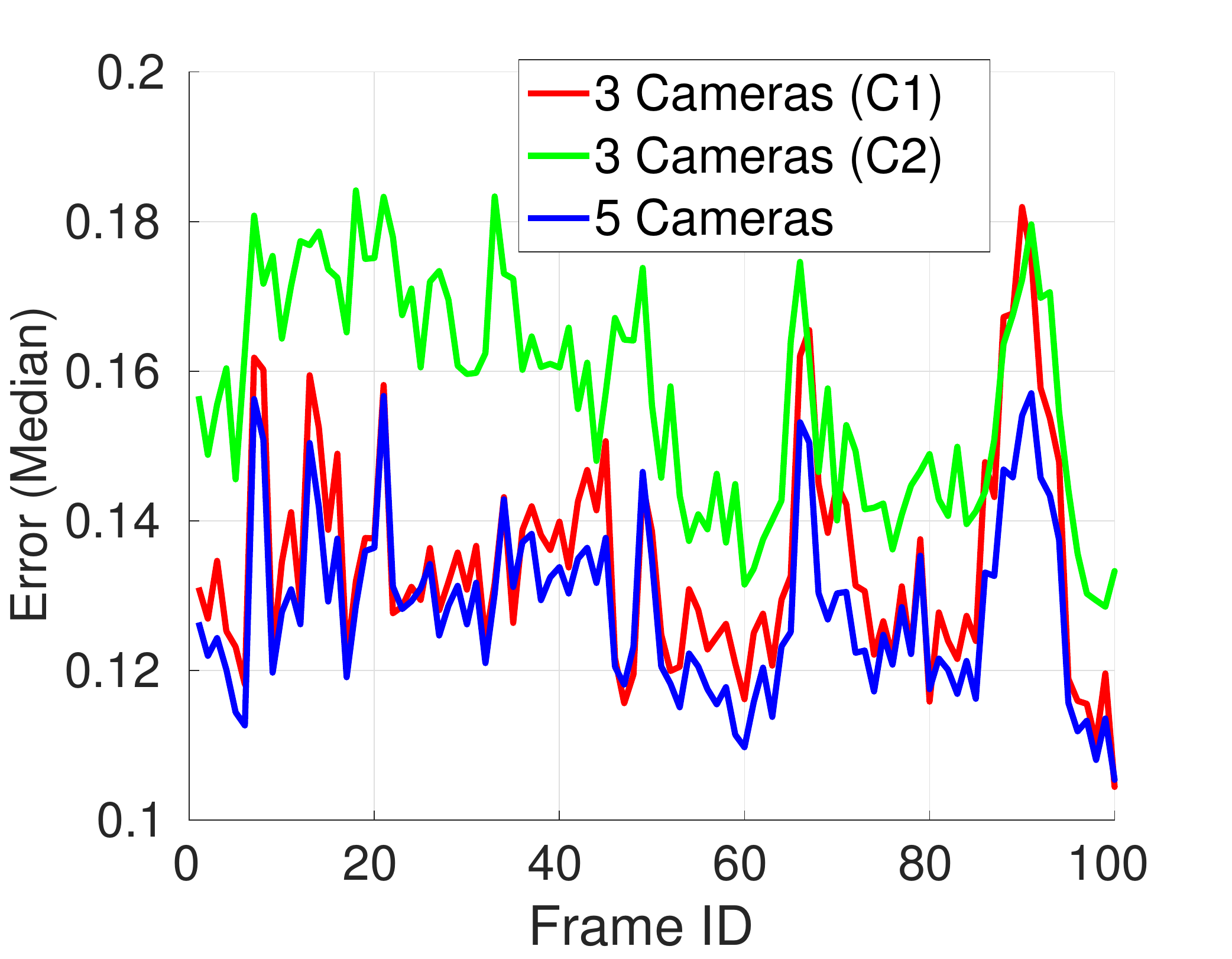} \vspace{-0.3em}\\
		\vspace{-0.5em}
		\small{(a) \em{South Buona Vista}} & \small{(b) \em{One North}}\\
	\end{tabular}
	\caption{\fvzh{Performance with different camera configurations. ``3 Cameras (C1)" means using leftmost, central and rightmost cameras. ``3 Cameras (C2)" means using three central cameras.}}
	\label{fig:camconfiguration}
	\vspace{-1.0em}
\end{figure}

We also evaluate the performance of the different filters. 
As can be seen in 
\figref{fig:effect_filtering}, the best cost filter mainly helps with filtering out unreliable estimates in textureless areas, e.g., the sky and building facade, thus greatly reducing the error. The uniqueness ratio filter further helps to remove unreliable estimates in the areas with repetitive patterns, e.g., the road. The local consistency filter mainly discards inconsistent estimates. 
In contrast to the other filters, applying the local consistency filter leads to a smaller improvement. 
\fvzh{From \figref{fig:medianErrFilters}, we can see that the median and mean values of the errors are finally decreased by more than 40$\%$ and 60$\%$ with these filters.} 
This will greatly improve the TSDF-based fusion result.

At last, \fvzh{we evaluate the performance with different numbers of cameras as shown in \figref{fig:camconfiguration}. Though the configurations with 3 cameras save about 17$\%$ of run time, they have higher errors than that with 5 cameras, which indicates that more cameras should be adopted for better depth estimation.}

\begin{figure}[t]
	\centering
	\begin{tabular}{*{2}{c@{\hspace{5px}}}}
		\includegraphics[trim=0cm 1.4cm 0cm 0cm, clip,width=0.22\textwidth]{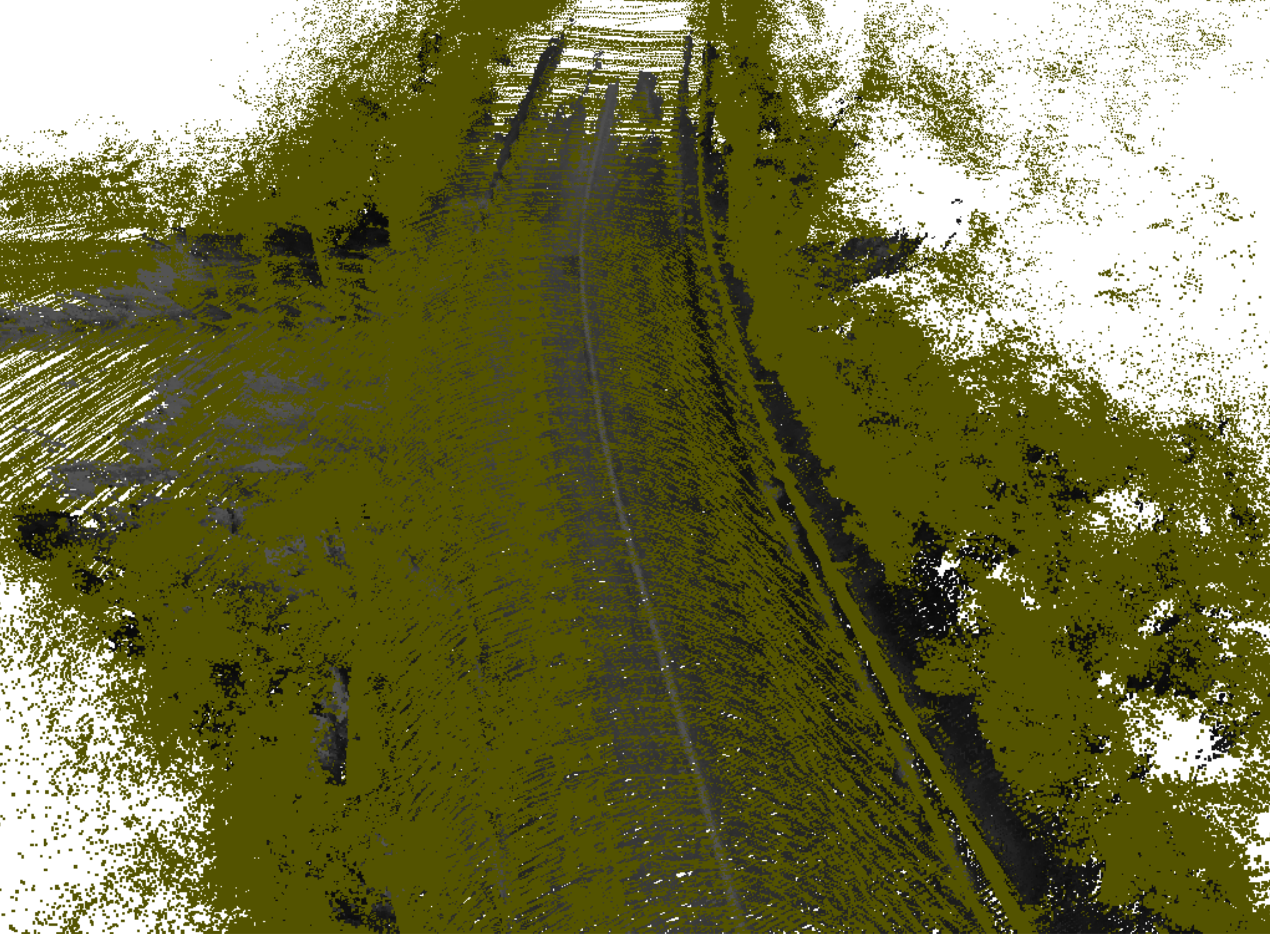} &
		\includegraphics[trim=0cm 1cm 0cm 0.35cm, clip,  width=0.22\textwidth]{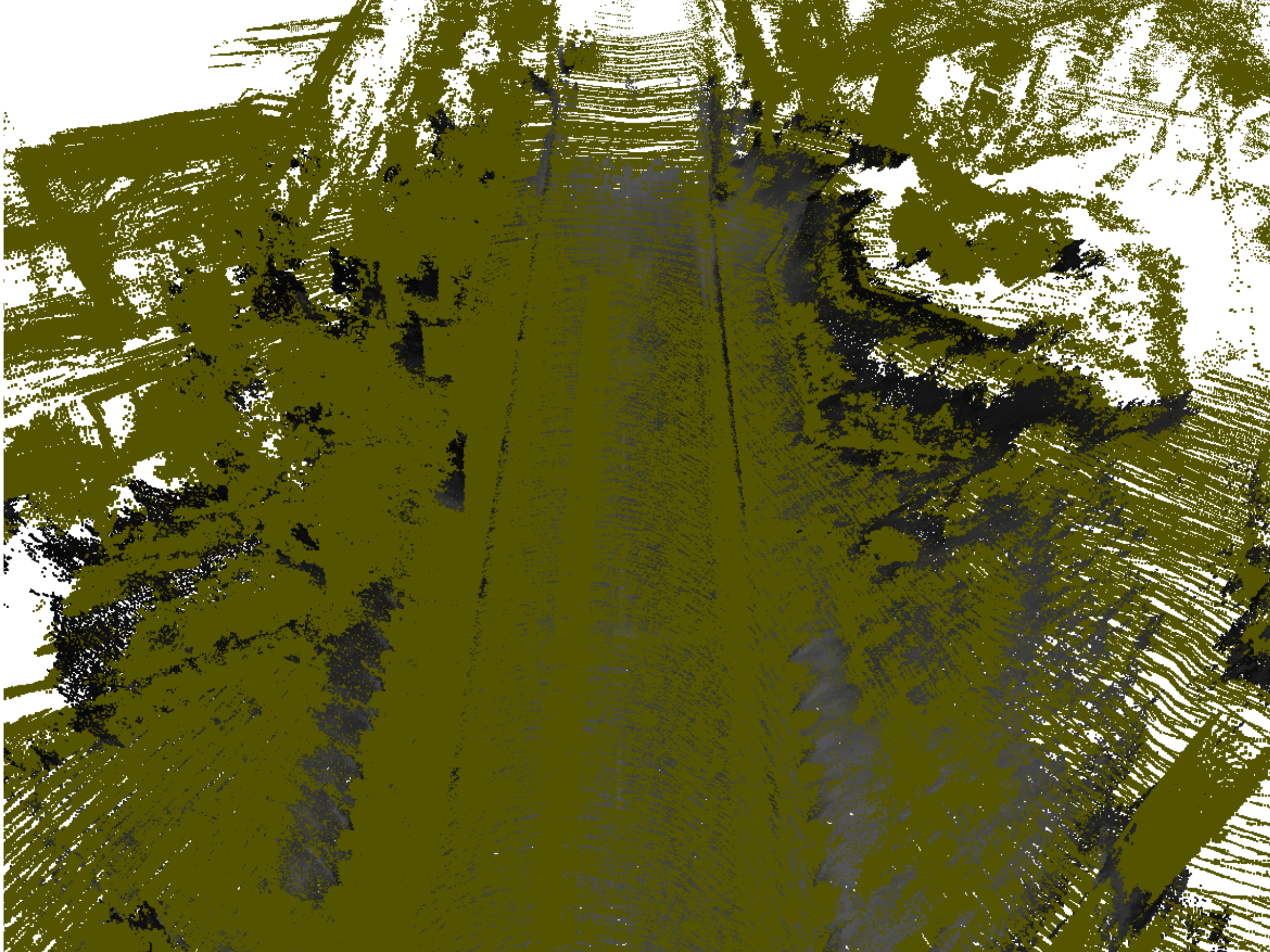} \vspace{-0.3em}\\
		\small{(a) \em{South Buona Vista}} & \small{(b) \em{One North}}\\
		\vspace{-1.5em}
	\end{tabular}
	\caption{Local map for a certain frame. The green points are the accumulated LiDAR data.}
	\vspace{-2.0em}
	\label{fig:reconstruction}
\end{figure}

\subsection{Evaluation of the 3D Mapping Stage}
To evaluate the quality of the 3D maps built by our approach, we compare maps built using our depth maps with maps constructed using the ground truth depth maps. 
We use accuracy and completeness as our evaluation criteria~\cite{schops2017multi}. 

\begin{figure}[t]
	\centering
	\begin{tabular}{*{2}{c@{\hspace{0px}}}}
	\hspace{-10px}	\includegraphics[width=0.26\textwidth]{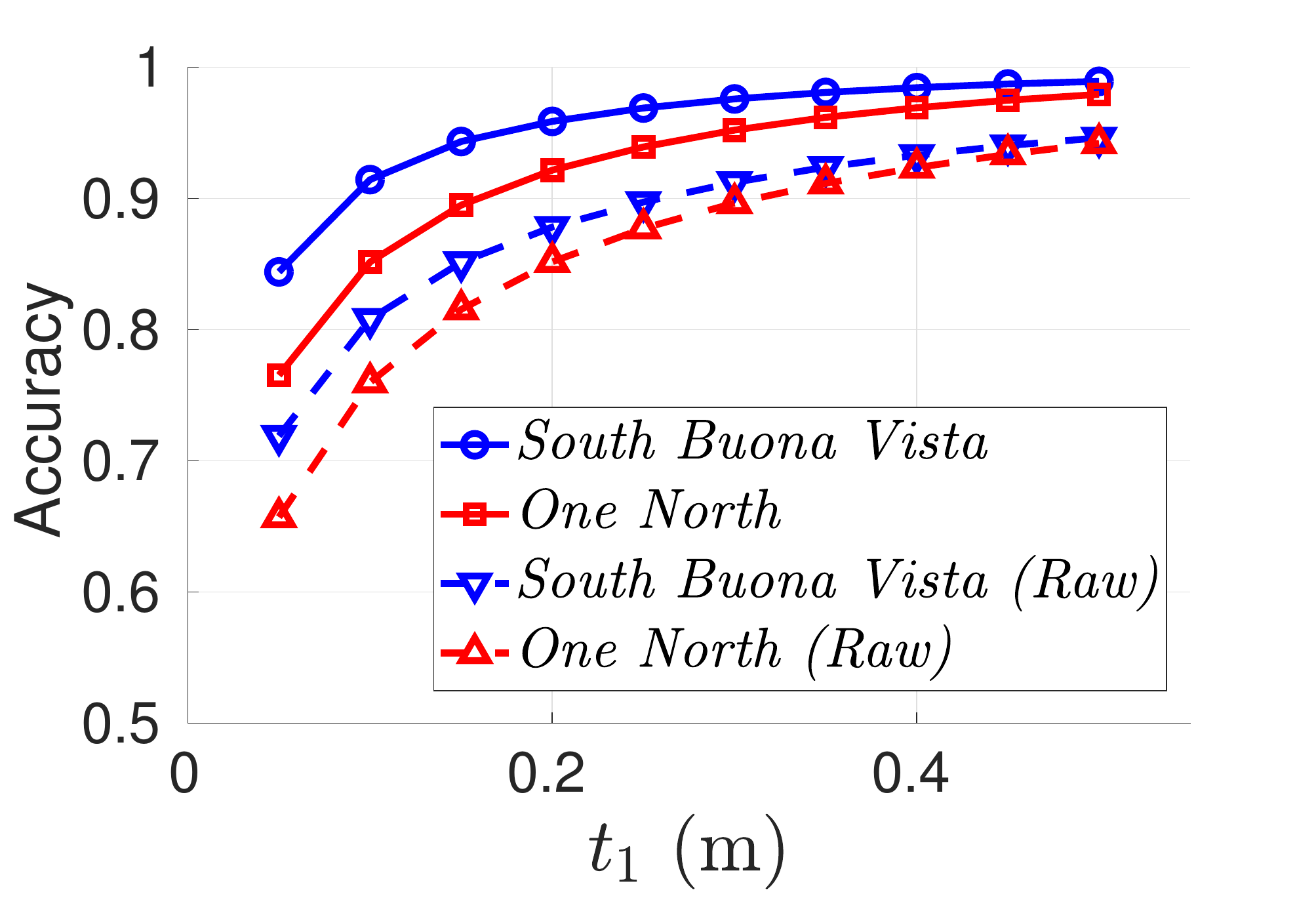} &
	\hspace{-10px}	\includegraphics[width=0.26\textwidth]{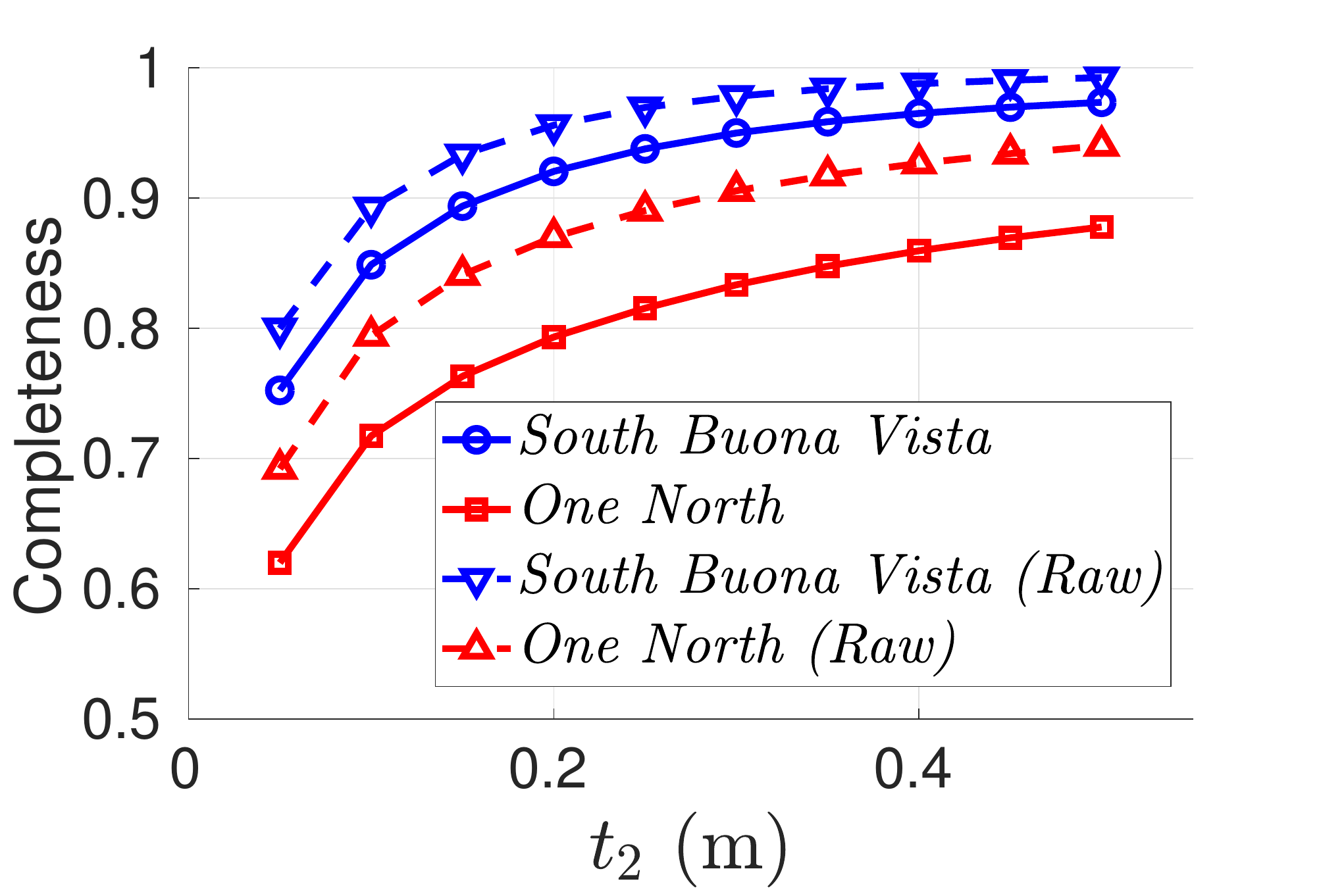} \\
	\end{tabular}
	\vspace{-0.8em}
	\caption{Evaluation of the mapping results with and without (raw) unreliable voxels removal.}
	\vspace{-1.em}
	\label{fig:reconstructionError}
\end{figure}

Given two sets of point clouds $S$ and $S'$, for each point $p$ in $S$, we define $d_p(S, S')$ as the distance of $p$ to its closest point $p'$ in $S'$. Let the 3D point sets computed from ground-truth depth maps and computed depth maps be $S_{gt}$ and $S_{c}$ respectively, then the accuracy can be defined as the ratio of the points in $S_c$ with $d_p(S_c, S_{gt}) \leq t_1$ out of all points in $S_c$. Similarly,  the completeness is defined as the ratio of the points in $S_{gt}$ with $d_p(S_{gt}, S_c) \leq t_2$ out of all points in $S_{gt}$. $t_1$ and $t_2$ are the tolerance parameters. We experimented with the same frames used in \secref{sec:evaluation_depth}, and the results are shown in \figref{fig:reconstructionError}. We can see that for both datasets, more than $85\%$ of the reconstructed points have an error of less than 0.1m. We find the performance in terms of accuracy to be good considering that we use a voxel size of 0.05m. In terms of completeness, {\em South Buona Vista} has better performance than {\em One North}. This is because the environment of  {\em One North} is more complex with more occlusions than {\em South Buona Vista}, and thus, more unreliable depth information is filtered. However, we can see that even for  {\em One North}, the completeness is more than $80\%$ with a threshold of 0.25m. \zh{\figref{fig:reconstructionError} also shows the result without removing voxels that do not have a sufficient number of observations. We can see that this improves the completeness at the expense of accuracy.}

\figref{fig:reconstruction} shows the computed 3D point clouds and accumulated LiDAR data. We can see that the recovered 3D map is well aligned with the LiDAR data. 
The figure also shows that the incompleteness mainly occurs in areas far away from the current vehicle position. This is consistent with our analysis of depth estimation.

\subsection{Evaluation of object detection}
\begin{figure}[t]
	\centering
	\begin{tabular}{*{2}{c@{\hspace{10px}}}}
		\hspace{-10px}	\includegraphics[trim=0cm 3cm 0cm 0cm, clip,width=0.2\textwidth]{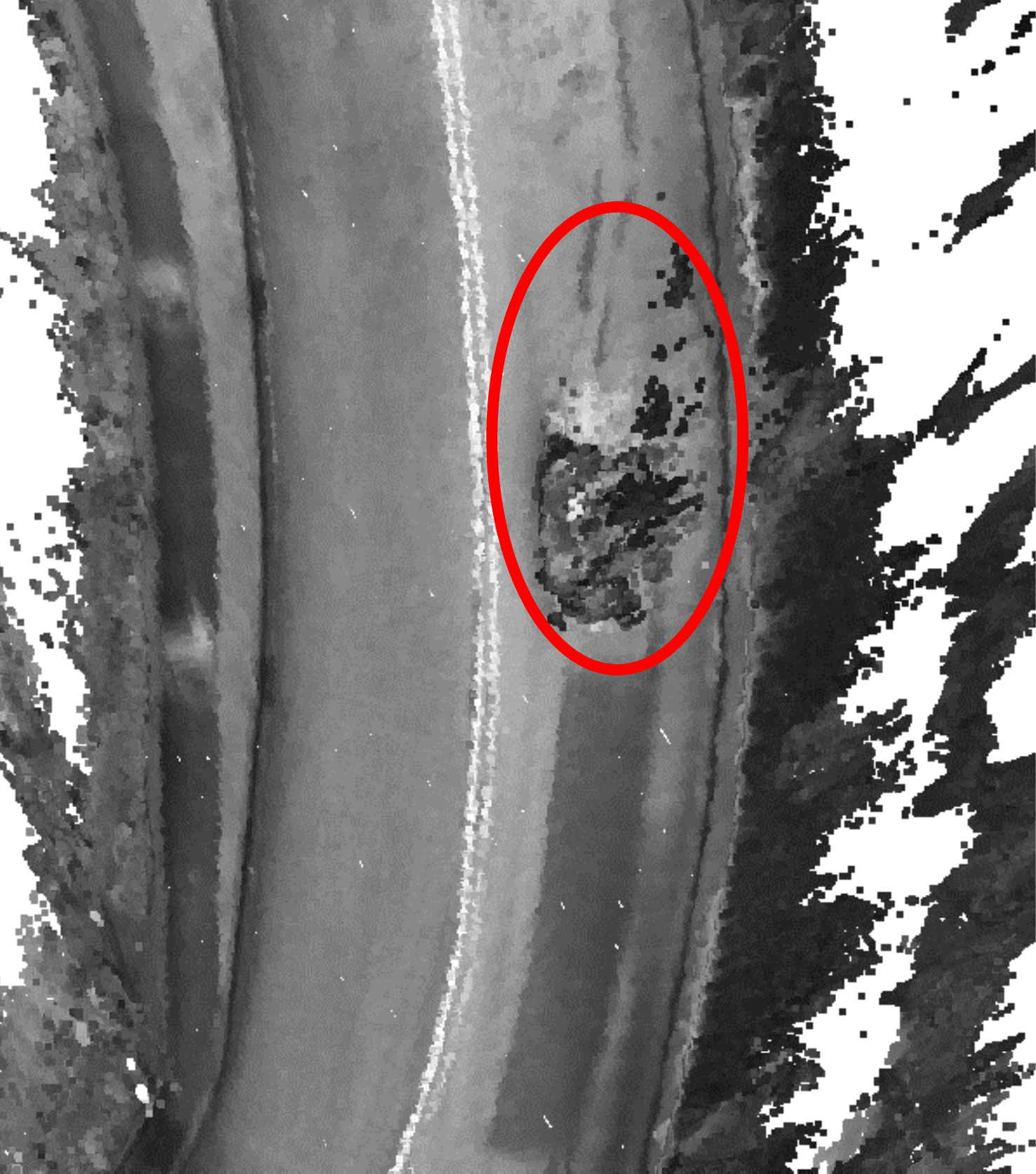} &
		\includegraphics[trim=0cm 3cm 0cm 0cm, clip, width=0.2\textwidth]{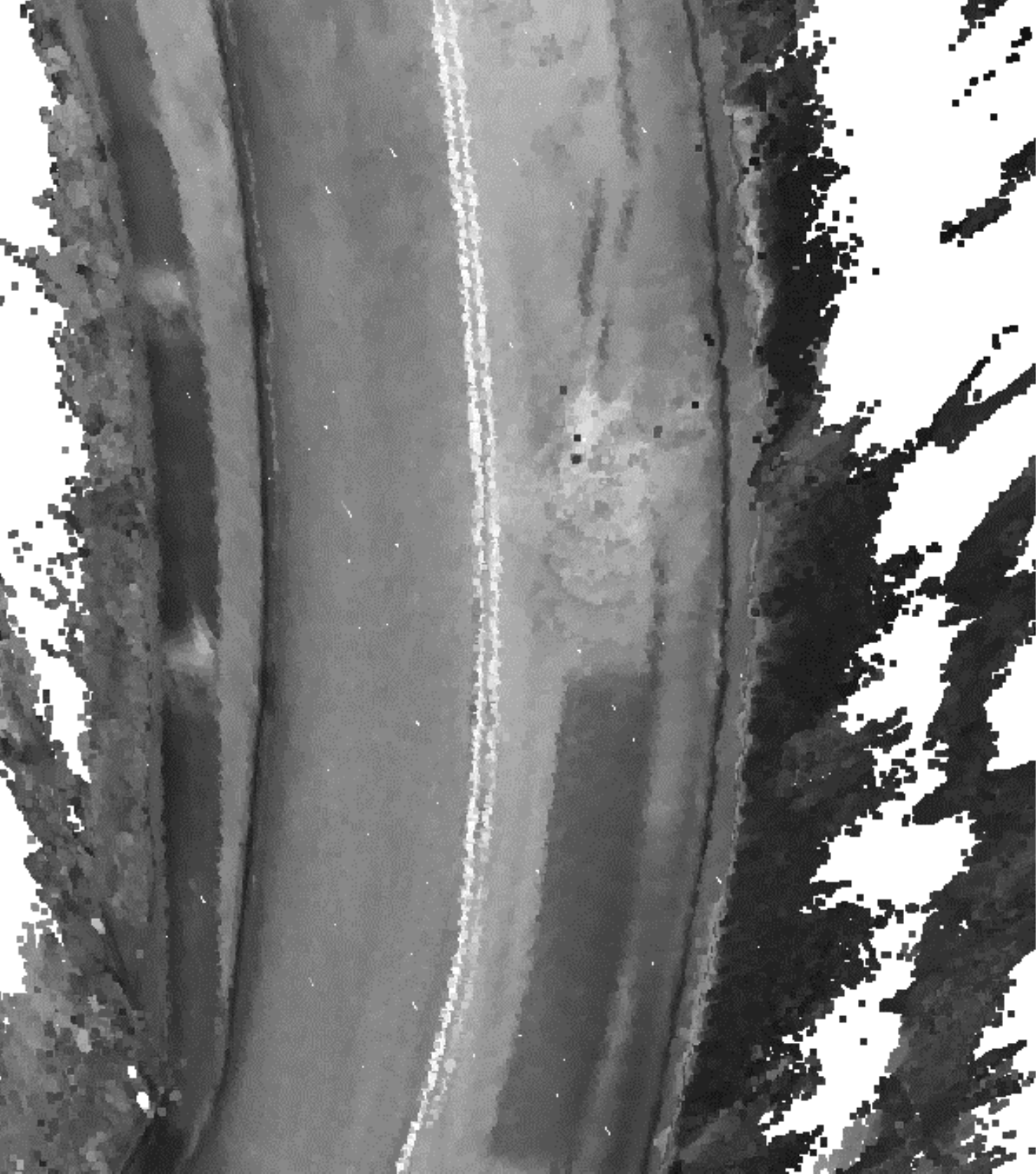} \\
		\vspace{-1.5em}
	\end{tabular}
	\caption{Example of reconstructed 3D points (left) without and (right) with moving object detection for {\em South Buona Vista}.}
	\vspace{-1.8em}
	\label{fig:effectObjectDetection}
\end{figure}

\figref{fig:effectObjectDetection} shows the mapping result for a certain frame without and  with the moving object detection. 
We can see that even though we prune unreliable voxels using the observation weights, there is still a trail of points caused by a moving object in the map. With moving object detection, the moving vehicle is completely filtered out in the map. We find that the object bounding box does not impact the mapping results, although the bounding box is not as tight as an object mask and may cause some background information to be filtered out. With temporal integration, the missing parts of the map and around the moving vehicle can be recovered.

\section{Conclusions}

In this paper, we have proposed a real-time dense mapping method for self-driving vehicles purely based on fisheye cameras. In order to achieve both depth map accuracy and run-time efficiency, we proposed a novel multi-scale depth map estimation strategy. To filter out the noisy depth estimation in featureless areas, we evaluated the impact of several depth filters. In order to handle  moving objects during TSDF fusion, we adopted a fast one-stage neural network for object detection and fine-tuned it on our labeled fisheye images. To make the whole system scalable to large scenes, we used a swapping strategy based on 3D location information. The experimental results demonstrate that our whole pipeline can achieve good accuracy and reasonable completeness compared to LiDAR data while running in real-time on the vehicle. 
\fvzh{We plan to study the effect of integrating higher-level scene understanding into the depth estimation and evaluate with different weather conditions in the future.}

\bibliographystyle{abbrvnat}
\footnotesize{
\bibliography{main}
}
\end{document}